\PassOptionsToPackage{table}{xcolor}
\documentclass[11pt, a4paper, logo, nonumbering]{style}
\usepackage{dblfloatfix}
\usepackage{caption}
\usepackage{float}
\usepackage{xspace}
\usepackage{pifont}
\usepackage{multirow}
\usepackage{tcolorbox}
\tcbuselibrary{breakable}
\usepackage{longtable}
\usepackage{hyperref}
\usepackage{amsfonts}
\usepackage{amsmath}
\usepackage{amssymb}
\usepackage{adjustbox}

\usepackage[bottom]{footmisc}
\usepackage{listings}
\usepackage{setspace}

\usepackage{array}
\usepackage{tabularx}
\usepackage{booktabs}

\usepackage{graphicx}
\usepackage{wrapfig}
\usepackage{tikz}
\usetikzlibrary{arrows.meta, positioning, shapes.geometric, fit, backgrounds}
\usepackage{pgfplots}
\pgfplotsset{compat=1.18}
\definecolor{oursblue}{HTML}{2C6FB5}
\definecolor{basegray}{HTML}{7A7F87}
\definecolor{viz1}{HTML}{2A78D6}
\definecolor{viz2}{HTML}{1BAF7A}
\definecolor{viz3}{HTML}{EDA100}
\definecolor{viz4}{HTML}{008300}
\definecolor{viz5}{HTML}{4A3AA7}
\definecolor{vizmine}{HTML}{EB6834}

\definecolor{badred}{RGB}{163,35,50}
\definecolor{goodgreen}{RGB}{19,116,67}
\definecolor{oursblue}{RGB}{55,115,170}
\definecolor{leakbadred}{RGB}{163,35,50}
\definecolor{leakgoodgreen}{RGB}{19,116,67}
\definecolor{leakoursblue}{RGB}{55,115,170}

\usepackage{multicol}

\definecolor{codebackground}{rgb}{0.95,0.95,0.95}
\definecolor{codeframe}{rgb}{0.8,0.8,0.8}

\lstdefinestyle{code}{
    backgroundcolor=\color{codebackground},
    frame=single,
    rulecolor=\color{codeframe},
    basicstyle=\ttfamily\footnotesize,
    keywordstyle=\bfseries,
    showspaces=false,
    showstringspaces=false,
    showtabs=false,
    tabsize=4,
    captionpos=b,
    breaklines=true,
    breakatwhitespace=true,
    sensitive=true,
    breakindent=0pt,
}

\makeatletter
\def\@BTrule[#1]{%
  \ifx\longtable\undefined
    \let\@BTswitch\@BTnormal
  \else\ifx\hline\LT@hline
    \nobreak
    \let\@BTswitch\@BLTrule
  \else
     \let\@BTswitch\@BTnormal
  \fi\fi
  \global\@thisrulewidth=#1\relax
  \ifnum\@thisruleclass=\tw@\vskip\@aboverulesep\else
  \ifnum\@lastruleclass=\z@\vskip\@aboverulesep\else
  \ifnum\@lastruleclass=\@ne\vskip\doublerulesep\fi\fi\fi
  \@BTswitch}
\makeatother

\addto\extrasenglish{
}



\definecolor{absframe}{HTML}{ED6000}
\makeatletter

\newcommand{\teaserfigure}{%
  \begin{figure}[!t]
    \centering
    \includegraphics[width=\linewidth]{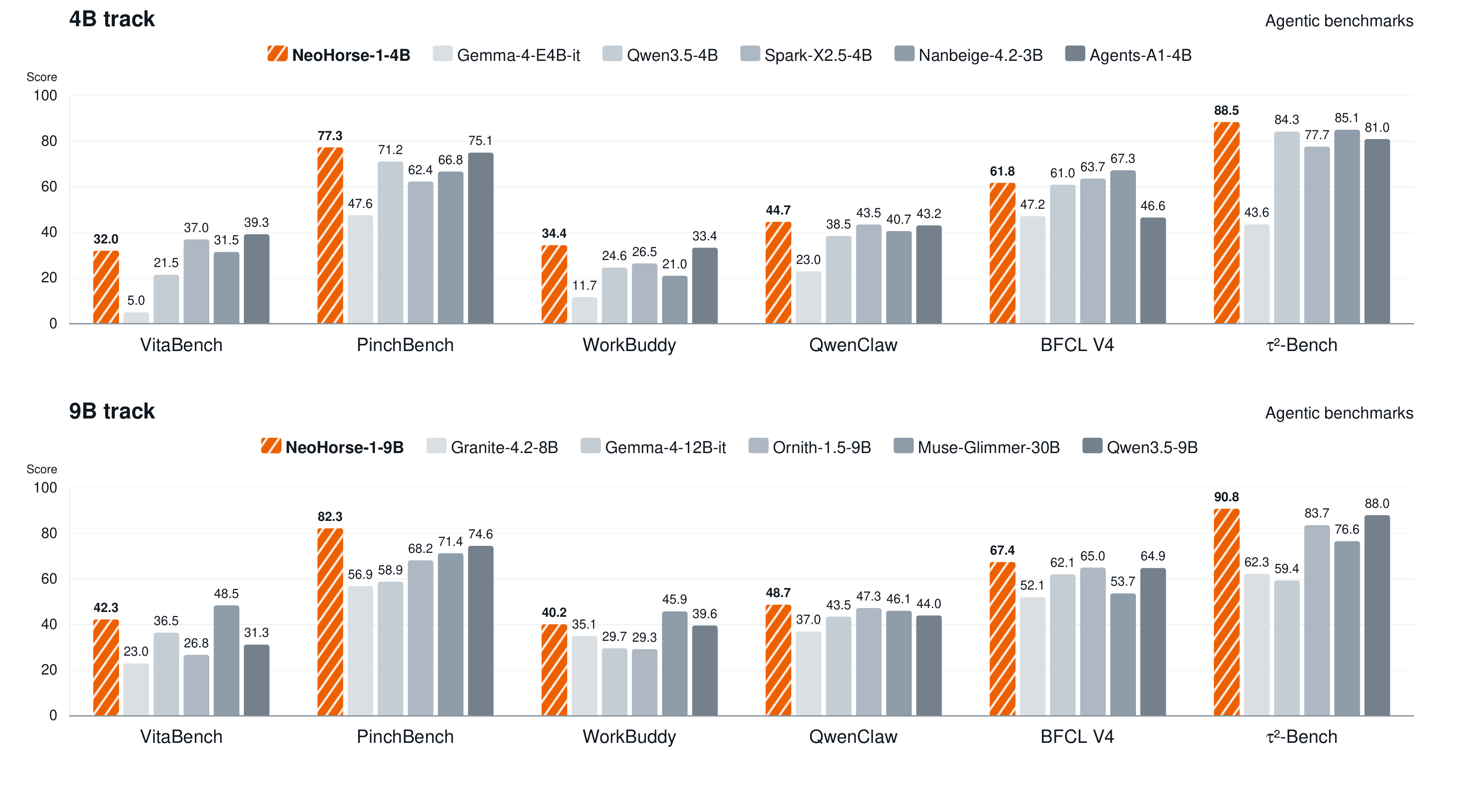}
    \captionsetup{font=small,skip=4pt}
    \caption{\textbf{Comparison on six agentic benchmarks in the 4B and 9B tracks.}
    Orange bars denote \modelname{}-4B (top) and \modelname{}-9B (bottom);
    gray bars denote the comparison models in each track, with model sizes
    indicated in the legends. Full results are reported in
    Tables~\ref{tab:results-4b} and~\ref{tab:results-large}.}
    \label{fig:headline-results}
  \end{figure}
}

\renewcommand{\abscontent}{%
  \noindent
  \begin{tcolorbox}[breakable, colframe=absframe, colback=absframe!5, boxrule=1pt, arc=8pt,
                    left=10pt, right=10pt, top=10pt, bottom=10pt]
    \centerline{\fontsize{15pt}{14pt}\selectfont\textbf{Abstract}}\vspace{2ex}
    {\absfont \theabstract}%
    \@ifundefined{@keywords}{}{%
      \vskip1em \noindent \keywordsfont Keywords: \@keywords}%
  \end{tcolorbox}%
}
\makeatother

\makeatletter
\renewcommand{\maketitle}{\bgroup\setlength{\parindent}{0pt}
        \vspace*{3pt}
	\begin{adjustwidth}{0pt}{0pt}
		\begin{flushleft}
			{
				{\raggedright \titlefont \@title\par}%
				\vskip20pt
				{\raggedright \@author\par}
				\vskip30pt %
			}%
		\end{flushleft}
	\end{adjustwidth}
	\egroup
	\thispagestyle{firststyle}%
	{%
		{\abscontent}
	}%
}
\makeatother

\newcommand{\modelname}{\emph{NeoHorse-1}}

\title{\centering{\raisebox{-0.12em}{\includegraphics[height=1.0em]{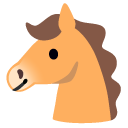}} \modelname: Towards Recursive Self-Improvement via Agentic Post-Training with Routing Harness}}

\author[]{
\setlength{\parskip}{0pt}%
{\normalfont\fontsize{11}{13}\selectfont
NeoHorse Team\par}
\vspace{6pt}
{\small
\href{https://hf.co/collections/TokenRhythm/neohorse-1}{\raisebox{-1.5pt}{\includegraphics[height=1.05em]{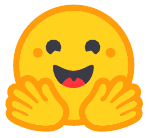}}\hspace{0.4em}\textcolor{absframe}{\texttt{https://hf.co/collections/TokenRhythm/neohorse-1}}}\par
\vspace{1pt}
\href{https://github.com/TokenRhythm/NeoHorse}{\raisebox{-1.5pt}{\includegraphics[height=1.05em]{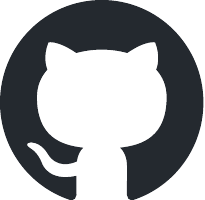}}\hspace{0.4em}\textcolor{absframe}{\texttt{https://github.com/TokenRhythm/NeoHorse}}}\par}
}

\renewcommand{\phi}{\varphi}

\renewcommand{\geq}{\geqslant}

\renewcommand{\epsilon}{\varepsilon}
\renewcommand{\imath}{\mathrm{i}}

\newlength{\restsubwidth}
\newlength{\restsubheight}
\newlength{\restsubmoreheight}
\newcommand{\rest}[2]{%
        \settowidth{\restsubwidth}{\ensuremath{#2}}
        \settoheight{\restsubheight}{\ensuremath{{}_{#2}}}
        \ensuremath{{#1\hskip 0.5pt}_{\vrule\kern2pt\parbox[b][%
        4pt][b]{\the\restsubwidth}{%
                        \ensuremath{{}_{#2}}}}}
        }

\begin{abstract}
  Recursive self-improvement (RSI) requires a concrete mechanism through which an AI system observes its own capabilities and converts that
  evidence into the next round of learning. We argue that a deployed
  routing harness already contains such a mechanism: beyond task outputs,
  agentic interaction leaves execution trajectories together with
  observable evidence of what a model can and cannot yet do. We present
  \modelname{}, a family of agent-native models developed to explore
  this path through agentic post-training. Our system couples a
  heterogeneous model pool with intelligent routing, which records, for
  every turn, the capability demand predicted, the service tier selected, and
  the interaction that followed. These records are converted into
  user-turn training examples that preserve interleaved reasoning, tool
  calls, and harness context, and are admitted through structural
  validation, six-dimensional semantic evaluation, and subscene-level
  labeling. Routing signals provide estimates of capability demand:
  they organize supervised fine-tuning into a
  three-stage curriculum and extend naturally to routing-guided on-policy
  distillation, where a teacher supervises student-generated responses
  under the same staged progression. Finally, a capability-guided
  allocation step turns evaluation feedback into the next training
  mixture, closing an evaluation--selection--update loop in which what the
  system learns to do shapes what it learns from next. Across ten
  benchmarks spanning harness-based agents, tool use, coding, and
  instruction following, post-training lifts the macro-average score of
  the 4B model from 58.94 to 64.87 and of the 9B model from 65.60 to
  69.04, substantially narrowing the aggregate gap between the
  post-trained 4B model and the 9B base model. We further adapt NeoHorse-1-4B
  into NeoHorse-Jev-4B, a prefill-only model that returns structured decisions
  and probabilities for agent workflows. \modelname{} constitutes an
  initial prototype of this feedback-driven process, and we outline how
  sustaining it across iterations can move harness-mediated RSI from
  design to practice.
\end{abstract}

\begin{document}
\emergencystretch=3em
\tolerance=2000
\maketitle

\teaserfigure

\section{Introduction}\label{sec:intro}
\begin{center}
\textit{``Distance tests a horse's stamina; Time reveals a man's heart.''} \small{--- Chinese idiom}
\end{center}

Recursive self-improvement (RSI) describes a broad direction in which AI systems take a growing part in the process of their own improvement---from refining individual responses and reshaping their execution harness, to learning from self-generated experience and, in an emerging line of work, automating parts of AI research itself~\citep{chen2026recursive,rsi202608}. Its appeal is structural: once model improvement itself becomes partially automated, each generation can contribute to producing the next, turning isolated training efforts into a compounding process that is less bounded by manually curated data and human supervision. Realizing this vision, however, requires a concrete mechanism through which a system observes its own capabilities and converts that evidence into the next round of learning. Agents are natural carriers of such a mechanism. When an agent writes code, investigates a question, or operates software, it leaves a record of its decisions, tool interactions, and task outcomes; such interaction trajectories and executable tasks have already been used to train agentic models~\citep{zeng2023agenttuning,chen2024agentflan,song2024agentbank,qwen2026codernext}. The opportunity extends beyond treating these records as static supervision: agentic interaction produces both task experience to learn from and observable evidence of model strengths and limitations, which can shape what a model learns next---precisely the feedback that RSI requires. We introduce \modelname{}, a family of agent-native models developed to explore this path through routing-guided agentic training.
Figure~\ref{fig:headline-results} provides an overview of the agentic benchmark results at both model scales.

A harness is the execution layer that manages an agent's context, tools, and interaction with its environment~\citep{yang2024sweagent,kim2026interplay}. Adding agentic routing allows this layer to select models according to the request and the evolving interaction state~\citep{chen2023frugalgpt,ong2024routellm}. Building on the harness-native data flywheel of Agentic Routing~\citep{liu2026agenticrouting}, our system design combines a heterogeneous model pool with multiple harnesses, including OpenSquilla~\citep{opensquilla2026}, across diverse real-world tasks. This design allows training experience to span different model behaviors and execution environments. Routing records link estimated capability demand, the model actually used, and the subsequent interaction, making them useful for organizing training experience.

\begin{figure}[t]
  \centering
  \includegraphics[width=\linewidth]{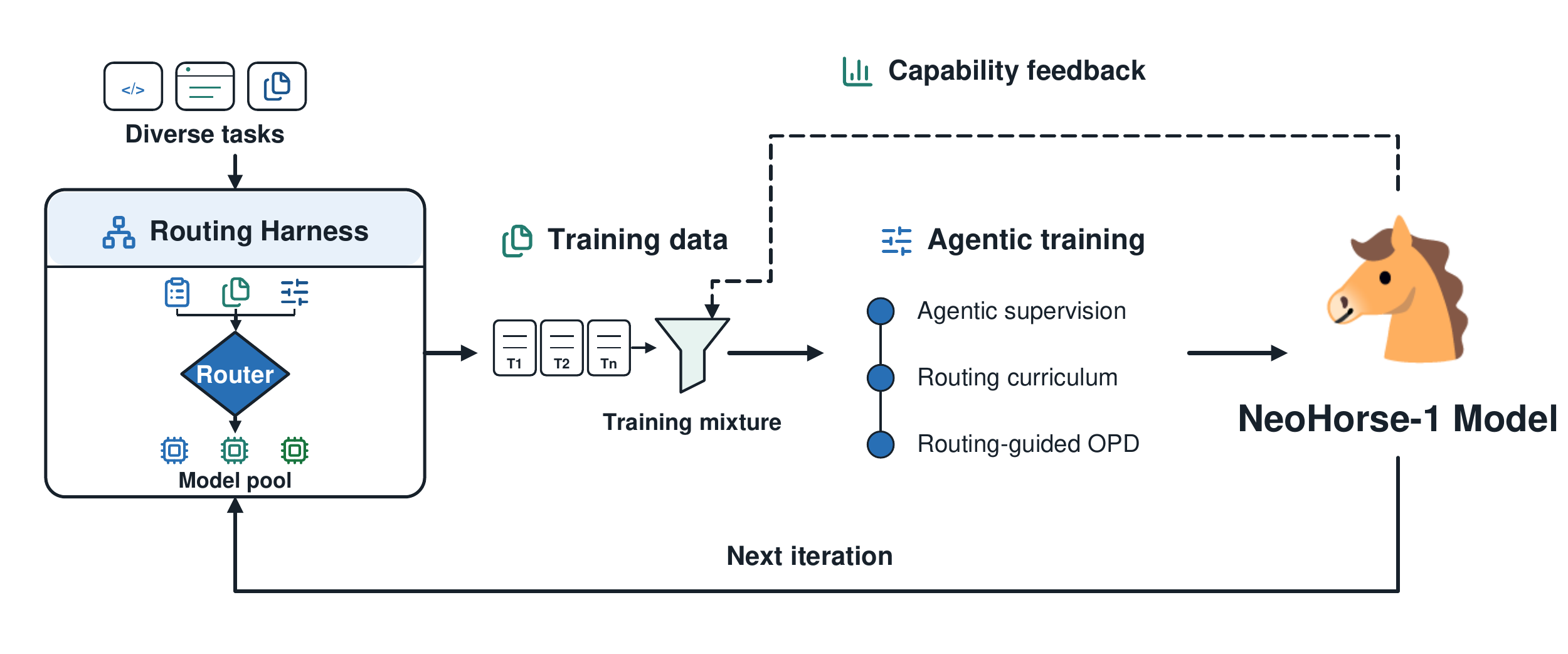}
  \caption{\textbf{Towards RSI through routing-guided agentic training.}
  Diverse tasks generate interaction experience through routing harnesses
  backed by heterogeneous model pools. This experience is organized into a training mixture
  for \modelname{}. Capability feedback guides the next training distribution,
  while updated models return to the harness for subsequent iterations. The agentic training stack
  summarizes the post-training methods described in this report.}
  \label{fig:intro-overview}
\end{figure}

In \modelname{}, this path towards RSI centers on an evolving training distribution. Capability-level feedback guides the allocation of data for subsequent model updates, while continued harness execution supplies new interaction experience. When updated models return to the harness, their behavior reveals a new pattern of strengths and limitations that can inform the next round of training. In this loop, what the system learns to do influences what it learns from next. Because the harness can also draw on other models, the process connects multi-model experience with the continued improvement of individual models. Figure~\ref{fig:intro-overview} summarizes this loop.

\modelname{}'s post-training methods provide the learning component of this design. We first convert recorded interactions into user-turn training examples that preserve the historical and harness context under which each assistant response was produced, allowing the model to learn interleaved reasoning, tool use, and visible responses without detaching them from their execution conditions. Because agentic interactions vary substantially in capability demand, we use routing-derived scores to organize this experience as a curriculum~\citep{bengio2009curriculum,lee2024instruction}, progressively introducing higher-scored examples while retaining lower-scored coverage. SFT, however, still learns from recorded assistant responses, whereas deployment unfolds on the model's own generated prefixes. We therefore extend the same routing-guided progression to on-policy distillation~\citep{agarwal2024onpolicy}, where a teacher supervises responses generated by the student. Routing organizes the learning material, while on-policy supervision follows the student's evolving behavior. Together, these methods turn the varied experience of a routing system into capabilities within a unified model.

We study \modelname~at the 4B and 9B scales, with evaluation covering harness-based agents, instruction following, coding, tool use, and interactive tasks. Across the evaluation suite, agentic post-training raises the macro-average score from 58.94 to 64.87 at the 4B scale and from 65.60 to 69.04 at the 9B scale. The largest gains appear on harness-based and execution-intensive evaluations, while the post-trained 4B model substantially narrows the aggregate gap to the 9B base model.
Full results are reported in Tables~\ref{tab:results-4b} and~\ref{tab:results-large}.
Beyond response generation, we adapt NeoHorse-1-4B into NeoHorse-Jev-4B,
a prefill-only model for application-defined decisions and probabilities.
Section~\ref{sec:neohorse-jev} presents this adaptation and its text decision
evaluation, illustrating how the agentic backbone can support a focused
decision component within an existing workflow.
This report presents an initial model-training prototype on this path towards RSI. The next step is to extend this feedback-driven process across successive iterations and broader task settings, and to study whether its gains can be sustained as model capabilities evolve.

\section{Related Work}\label{sec:related}
\subsection{Agentic Model Post-Training}
\label{sec:rw-agentic-models}


Trajectory-based supervised fine-tuning (SFT) provides a practical route for
transferring planning and tool use into model parameters. FireAct and
AgentTuning learn from interaction trajectories, while Agent-FLAN and AgentBank
show that data composition and scale affect generalization
\cite{chen2023fireact,zeng2023agenttuning,chen2024agentflan,song2024agentbank}.
Llama 3 extends this recipe with synthetic multi-step tool-use data and
iterative SFT, rejection sampling, and direct preference optimization
\cite{grattafiori2024llama3}.

Trajectory SFT mainly imitates behavior from a fixed teacher policy.
On-policy distillation (OPD) reduces this distribution gap by letting the
student generate trajectories while the teacher supplies token-level logits on
student-visited states \cite{agarwal2024onpolicy,lu2025opd}. Compared with
conventional teacher-trajectory distillation
\cite{hinton2015distilling}, OPD provides denser process supervision aligned
with the student's evolving behavior.

The interaction harness determines which tools, observations, and feedback
enter the training distribution. SWE-agent and the Interplay of Harness Design
and Post-Training show that interface choices affect agent performance and
robustness, while Terminal-Lego highlights the value of explicitly structured,
environment-grounded trajectories
\cite{yang2024sweagent,kim2026interplay,yang2026terminaltrajectories}.
Agentic RL with executable environment feedback then replaces fixed
behavioral labels with rewards derived from tool execution and task
outcomes. Search-R1, ReTool, and RAGEN study this paradigm for search,
tool use, and long-horizon interaction, while Agent Lightning v1.0 keeps
the environment loop inside the deployment harness and Co-Harness jointly
updates the harness and model
\cite{jin2025searchr1,feng2025retool,wang2025ragen,
he2026agentlightning,chen2026coharness}.

\subsection{LLM Routing and Curriculum Learning}
\label{sec:rw-routing-curriculum}

LLM routing assigns each query to an appropriate model while balancing response quality against inference cost. FrugalGPT studies cost-aware model cascades~\cite{chen2023frugalgpt}, whereas RouteLLM learns to route queries between stronger and weaker models using preference data~\cite{ong2024routellm}. Agentic routing~\cite{liu2026agenticrouting} extends this idea to multi-agent LLM systems, where a decision layer selects the model or sub-agent best suited to handle each incoming request.

Curriculum learning is a training strategy that presents examples according to an estimated notion of difficulty~\cite{bengio2009curriculum}. It has been shown to be effective in LLM post-training~\cite{xie2025logic,lee2024instruction}. Existing approaches, however, often rely on explicit difficulty labels or dataset-specific heuristics, which can be costly or impractical to obtain. Routing systems provide an alternative signal: their request- and context-conditioned predictions estimate the relative capability demand of an interaction. We use this predicted demand to order examples for curriculum training, rather than treating the identity of the model ultimately served as a difficulty label, since the executed route may also reflect user overrides, service availability, and deployment policy.

\subsection{Recursive Self-Improvement}
\label{sec:rw-rsi}

Recursive self-improvement (RSI) denotes an iterative process in which an AI system uses experience, evaluations, or generated artifacts to improve its model, scaffold, or improvement procedure
\citep{good1966speculations,yudkowsky2008recursive}. Recent RSI research considers both what is improved—from agent behavior and
policy to the surrounding scaffold and the training or research process—and how
tightly generation, evaluation, and updating are linked within the resulting
feedback loop
\citep{chen2026recursive,ren2026selfimproving}. Across these targets,
system-level efforts have begun to automate components such as harness design,
serving infrastructure, and training pipelines
\citep{weng2026harness,claude2026,astra2026}.

Recent studies examine recursive improvement at the levels of
task behavior, agent scaffolds, and training or research procedures. MetaSkill-Evolve
jointly evolves task skills and the meta-skill that governs their improvement,
while AREX alternates evidence gathering with answer auditing
\citep{wang2026metaskillevolve,lu2026arex}. Self-Harness, Agentic Harness
Engineering, and Retrospective Harness Optimization use failures, observability,
and past trajectories to update harnesses
\citep{zhang2026selfharness,lin2026agenticharness,pan2026rho}. Continual Harness
extends this setting to reset-free online adaptation and model updates
\citep{karten2026continualharness}. At the training-process level, AI4AI-Bench
evaluates whether agents can modify training algorithms so that later runs
inherit improvements
\citep{chi2026ai4aibench}.

\section{Data from Routing Harness}\label{sec:data}

Post-training data for an agentic model is not adequately represented by static instruction--response pairs. It consists of execution trajectories that connect user requests, model reasoning, tool actions, environment observations, and task outcomes. Our data construction therefore centers on trajectories generated by the deployment harness, while public instruction, reasoning, tool-use, code, and preference data are used to broaden capability coverage. Consistent with recent agentic-model reports \citep{tongyi2025deepresearch,liu2026agenticrouting}, we preserve the execution context and observable outcome signals needed to learn not only final-answer generation, but also task progression, tool interaction, and recovery behavior. The remainder of this section describes the composition and serialization of the corpus, its quality control and labeling, the routing signals recorded by the harness, and finally how evaluation feedback reallocates subsequent training mixtures---the data-side groundwork for harness-mediated RSI.

\subsection{Data Composition}
\label{sec:data-composition}
We organize the corpus at three linked granularities. A \emph{trajectory} is a complete interaction executed by the deployment harness, preserving user requests, model responses, tool calls, environment observations, recovery attempts, and terminal outcomes. A \emph{user turn} begins with a user request and ends at the next user request or task termination; it serves as the basic serialized training unit. A \emph{subscene} groups adjacent user turns that share a local goal and thus spans one or more user requests; it serves as the unit for semantic characterization (Section~\ref{sec:data-labeling}). This organization connects full execution histories to learning examples and semantic units without breaking their provenance.

Within each user-turn record, the current request and its interleaved reasoning, tool calls, and observations are retained to preserve the reasoning--action--feedback chain. Earlier visible responses and tool interactions remain available as context, whereas reasoning from earlier turns is omitted. Each record remains linked to its parent trajectory and subscene, allowing quality, semantic, routing, and outcome signals to be aligned at their appropriate granularity. Related approaches to organizing reasoning context in multi-turn data are discussed by \citet{deepseekai2025v32} and the Qwen team \citep{qwen2025multiturn}.

The primary corpus consists of on the order of $10^5$--$10^6$ harness-generated trajectories. We additionally use publicly available data to broaden coverage across instruction following and dialogue, reasoning, tool use and code, agent interaction, and preference learning \citep{oasst2,ayaDataset,openThoughts3,openR1Math,liu2024toolace,liu2024apigen,nemotronSwe,helpsteer2,ultrafeedbackCleaned}. Corpus scale is reported by the number of trajectories $N_{\mathrm{traj}}$ and tokens $N_{\mathrm{tok}}$ after unified serialization, deduplication, and tokenizer freezing, with the resulting statistics recorded in the training manifest.

\subsection{Data Quality}
\label{sec:data-quality}

\paragraph{Deduplication and decontamination.}
The corpus is deduplicated at exact and near-duplicate granularity, and the same matching infrastructure screens every training candidate against our evaluation suites: records that overlap an evaluation item are removed from the training side, keeping the training corpus and the evaluation data disjoint (Section~\ref{sec:evaluation-guided-data}).

\paragraph{Structural validation.}
Each trajectory then undergoes rule-based structural validation. At the turn level, the pipeline reconstructs requests, model responses, tool calls, tool observations, and terminal events. It then verifies payload readability, supported message structure, request and response presence, causal event order, and closure of tool-call/result pairs through identifiers and execution branches. The same stage detects missing responses, orphan observations, duplicated or conflicting tool-call identifiers, unresolved internal calls, and ambiguous terminal branches. Because these properties are directly observable from the trajectory, they are evaluated using reproducible rules rather than model-based scores.

The structural gate produces three operational outcomes: internally complete, partially recoverable, and quarantined. Complete trajectories proceed directly to semantic evaluation; recoverable trajectories contribute only causally closed sub-trajectories; trajectories with ambiguous event ownership or no recoverable supervision target are quarantined. Structural validity establishes reliable serialization and replay, but does not imply correct tool selection or task success.

\paragraph{Semantic evaluation.}
For structurally usable trajectories, we construct a normalized semantic event stream and evaluate six independent quality dimensions: goal attainment, instruction adherence, tool use, evidence consistency, error recovery, and termination. These dimensions judge the quality of the execution and are distinct from the scene, goal, and outcome attributes used to characterize what the user asked for (Section~\ref{sec:data-labeling}). High-certainty failures---such as a missing final response, an unresolved tool call, or an unrecovered terminal error---are detected deterministically. Cases that require task-level interpretation are evaluated by a semantic judge that is restricted to evidence explicitly present in the trajectory. Every finding must be grounded in the corresponding events. Long trajectories are evaluated in segments and subsequently aggregated at the turn level so that intermediate failures can be distinguished from successful later recovery.

Each semantic dimension is assigned \texttt{PASS}, \texttt{WARN}, \texttt{FAIL}, or \texttt{NOT\_EVALUATED}, and evaluation coverage is stored separately. Missing evidence or an interrupted judge call is never converted into a positive verdict. The quality representation therefore retains the structural state, the six quality dimensions, and evidence coverage rather than compressing them into a single heuristic score. Training admission, review, and quarantine policies are defined over this structured representation.

\begin{figure}[t]
    \centering
    \includegraphics[width=\linewidth]{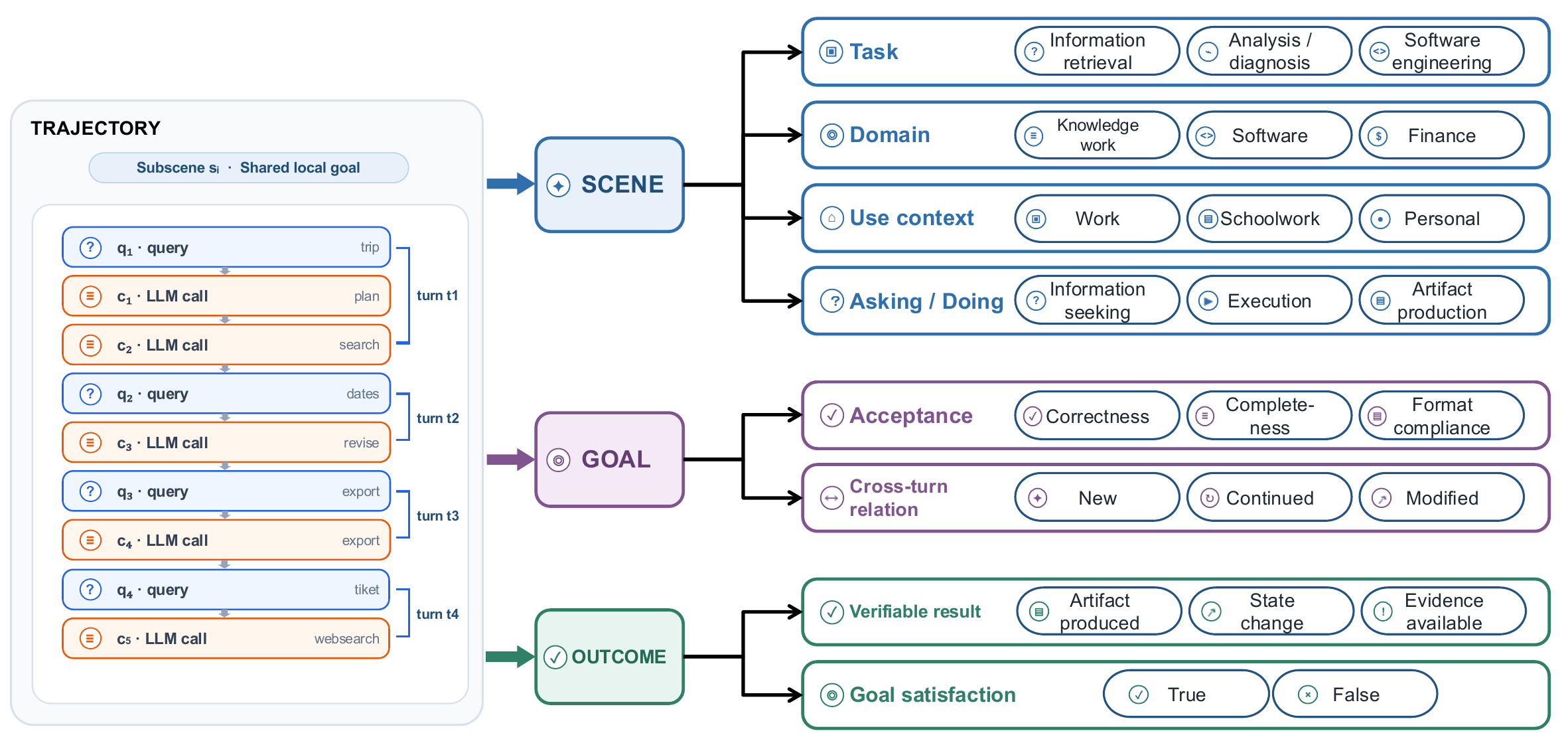}
    \caption{\textbf{Subscene-level scenario characterization.}
    A trajectory is represented as an ordered event stream of user queries and LLM calls; adjacent user turns that share a local goal form a subscene. The selected subscene is described through three complementary views---\emph{Scene}, \emph{Goal}, and \emph{Outcome}---with representative attributes shown on the right.}
    \label{fig:subscene-characterization}
\end{figure}

\subsection{Data Characterization}
\label{sec:data-labeling}

To support systematic improvements in user experience, we organize the attribute scheme around diverse usage scenarios. As illustrated in Fig.~\ref{fig:subscene-characterization}, the scheme characterizes each subscene along three axes: \emph{Scene} describes what the user is doing and in what context; \emph{Goal} states what the user expects to achieve and how success is to be judged; and \emph{Outcome} records the verifiable result of the attempt. Together, these axes connect user intent, agent execution, and outcome for capability analysis and data allocation.

Attributes are assigned at the \emph{subscene} level (Section~\ref{sec:data-composition}), which captures goal continuation, modification, interruption, and resumption within a conversation. On the Scene axis, closed taxonomies cover the task type and application domain, so that the corpus can be stratified by what the user is trying to do; each subscene receives one primary value and up to two secondary values. Use Context and Asking/Doing further characterize the setting and whether the request seeks information or execution. On the Goal axis, the user objective is decomposed into acceptance criteria that define how success is judged, and cross-turn relations mark whether a goal is new, continued, modified, resumed, or ambiguous. On the Outcome axis, the verifiable result of the attempt against the goal is recorded, so that the extent to which the task was actually satisfied can be distinguished from the process having run to completion.

To control noise from model-assisted annotation, each attribute retains its derivation method and confidence. Structural facts established by the source or deterministic rules cannot be overwritten by a semantic judge. Structural quality, turn-local reasoning policy, loss masks, and routing records (Section~\ref{sec:routing-data}) remain separate metadata. Together with the three axes, these signals localize capability gaps and guide subsequent data allocation.

\subsection{Agentic Routing Signals}
\label{sec:routing-data}

Together with the semantic attributes of Section~\ref{sec:data-labeling}, routing signals provide a complementary view of each trajectory. Scene, goal, and outcome attributes describe what the user requested and what the system achieved; the harness's routing module records the capability level predicted, selected, and actually served for each user turn. Aligning these fields yields a prediction--action--outcome record, allowing the corpus to be stratified jointly by user intent, service allocation, and observed result.

The harness router operates at the user-turn level and estimates capability demand from the current request, recent dialogue, previous routing decisions, and available execution state \citep{liu2026agenticrouting}. It assigns each turn to one of four service tiers: C0 handles bounded low-risk requests, C1 is the general-purpose default, C2 supports multi-step reasoning and execution, and C3 provides maximum capability or reliability. Policy controls may adjust this assignment in response to risk, context pressure, prior failure, or service constraints.

The tiers describe relative capability demand under the routing policy. Models, pricing, and inference configurations may change across deployments, and a C3 path may combine multiple proposers with an aggregator \citep{opensquilla2026}. Versioned tier semantics therefore keep routing records interpretable as the serving stack evolves.

For each turn, the corpus retains the router's raw prediction, the policy-adjusted decision, and the tier actually served, so that predicted demand, policy constraints, and executed action remain independently analyzable.

Each routing record is linked to its trajectory, so the corpus exposes completion, verification, and recovery patterns across capability-demand regions, while routing behavior is assessed using task completion, verification feedback, and recovery cost. This prediction--action--outcome separation \citep{liu2026agenticrouting} lets us isolate the routing estimate used for curriculum ordering in Section~\ref{sec:method-routing-curriculum} without treating the tier actually served as the training label. Outcome fields instead provide deficiency signals for the capability-guided allocation of Section~\ref{sec:evaluation-guided-data}. By feeding deployed routing outcomes into the next data mixture, the routing system functions both as a serving-time decision mechanism and as a source of data feedback within harness-mediated RSI.

\subsection{Capability-Guided Data Allocation}
\label{sec:evaluation-guided-data}
The quality dimensions of Section~\ref{sec:data-quality}, semantic attributes of Section~\ref{sec:data-labeling}, and routing signals of Section~\ref{sec:routing-data} organize deployment trajectories by user intent, capability demand, execution quality, and outcome. Routing places samples in capability-demand regions, while verification and outcome fields measure performance, yielding a stratification space over diverse tasks and usage scenarios.

At each iteration, the current model checkpoint is evaluated on a stratified suite kept disjoint from training by the decontamination screening of Section~\ref{sec:data-quality}. Results are aggregated across attributes, quality dimensions, outcome states, and routing tiers to form a model-deficiency profile. This profile shifts the next training mixture toward underperforming regions while preserving broad coverage. Verified successful trajectories provide positive supervision, while informative failures identify regions that require additional or rebalanced coverage in subsequent mixtures. These allocation decisions change the composition of the training data rather than introducing a separate failure-specific objective.

Between iterations, the harness continuously adds trajectories processed by the same quality, characterization, and routing pipeline. Existing and new data are reallocated together so that the corpus follows changes in usage patterns and system capability. When updated checkpoints return to the harness, subsequent trajectories reveal the next capability gaps, closing the evaluation--selection--update loop of harness-mediated RSI \citep{liu2026agenticrouting,longcat2025wowservice}. Within the Method section, routing-derived scores affect optimization through two scheduling mechanisms: they order user-turn examples in the three-stage SFT curriculum of Section~\ref{sec:method-routing-curriculum}, and the same progression schedules the starting contexts for on-policy distillation in Section~\ref{sec:method-opd}.

\section{Agentic Post-Training}\label{sec:method}

\subsection{Agentic Supervision}
\label{sec:method-agentic-supervision}

An agent trajectory records a model's interaction with an environment,
including user requests, model outputs, and tool results. In responding to a
user request, the assistant may reason, call a tool, and reason again after
receiving the tool result. This pattern is referred to as \emph{interleaved
thinking} \citep{anthropicThinking}. We use these records for SFT, supervising assistant responses within each user turn
and retaining earlier interactions as context.

\paragraph{User turns as training units.}
We define a \emph{user turn} as a user request together with the
assistant responses and tool interactions that follow it, up to the next
user request or the end of the recorded interaction. Tool results and
harness-injected messages do not initiate a new user turn. Each training
example contains one such turn and its historical context
(Section~\ref{sec:data-composition}). A turn may contain several assistant
responses interleaved with tool results; we supervise the retained assistant
target spans in a single sequence.

\paragraph{Serialization and context.}
We serialize each user-turn example using the Qwen3.5 chat template and
tool-call format \citep{qwen35ChatTemplate}. Consistent with its
reasoning-context convention,
we retain reasoning within the current turn when present and omit reasoning
from earlier turns. A similar context policy is described for DeepSeek-V3.2
\citep{deepseekai2025v32}. Earlier user requests, visible assistant responses,
tool calls, and tool results remain available as context.
We retain the recorded system instructions and harness-provided context so
that assistant targets remain paired with the conditions under which they
were produced.

\begin{figure}[t]
  \centering
  \includegraphics[width=\linewidth]{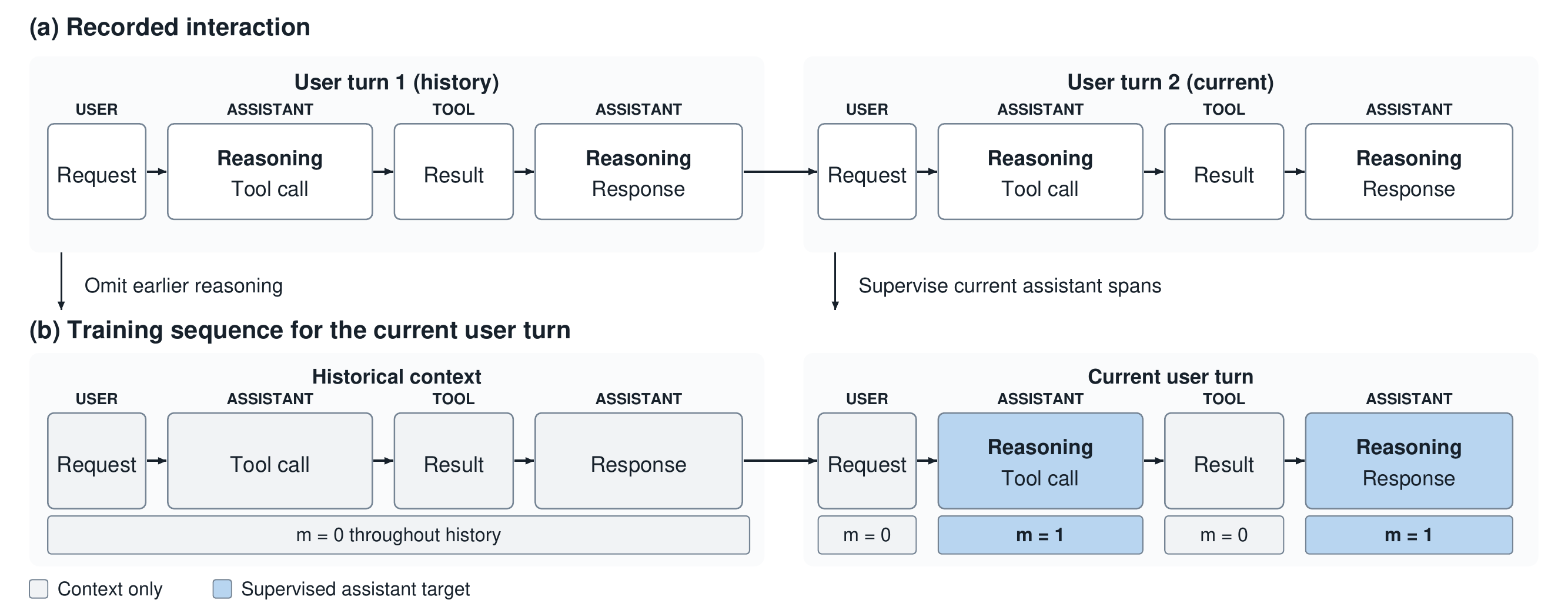}
  \caption{\textbf{Agentic supervision within a user turn.}
  A recorded interaction (top) is converted into a training sequence for
  the current user turn (bottom). Earlier reasoning is omitted, while visible
  responses and tool interactions remain as historical context. In the current
  user turn, assistant target spans receive prediction loss; user messages and
  tool results do not.
  Reasoning is retained when present, and assistant targets include
  end-of-response tokens. The sequence uses causal attention.
  System instructions, tool specifications, and retained harness messages
  are omitted from the schematic; they receive no prediction loss.
  Block widths do not reflect token counts.}
  \label{fig:agentic-supervision}
\end{figure}

Historical messages and all non-assistant spans---system instructions, tool
specifications, retained harness-provided context, user messages, and tool
results---receive no prediction loss. With causal attention, each
assistant response can use earlier actions and tool results in the sequence,
but not later ones. Figure~\ref{fig:agentic-supervision} illustrates this
construction.

\paragraph{SFT objective.}
Let $x_i=(x_{i,1},\ldots,x_{i,T_i})$ denote a serialized training sequence,
including its historical prefix and current turn. We use a binary token-level
loss mask $m_{i,t}$, set to one for tokens in the retained assistant target
spans of the current turn: reasoning when present, serialized tool calls and
their arguments, visible responses,
and end-of-response tokens. All other tokens, including padding, have
$m_{i,t}=0$. For a batch $\mathcal{B}$ of logical sequences, we minimize
\begin{equation}
  \mathcal{L}_{\mathrm{SFT}}(\theta;\mathcal{B})
  = -\frac{
      \displaystyle\sum_{i\in\mathcal{B}}\sum_{t=2}^{T_i}
      m_{i,t}\log p_{\theta}(x_{i,t}\mid x_{i,<t})
    }{
      \displaystyle\sum_{i\in\mathcal{B}}\sum_{t=2}^{T_i}m_{i,t}
    }.
  \label{eq:user-turn-masked-sft}
\end{equation}
This objective weights supervised tokens equally and normalizes by their
count across the batch, not by total sequence length or the number of turns.

\subsection{Routing-Guided Curriculum Learning}
\label{sec:method-routing-curriculum}

Agentic interactions vary in the capability they require, from routine
responses to complex planning and tool coordination. We use routing estimates
of this demand to organize SFT examples into a curriculum. Following the
C0--C3 capability ordering introduced in Section~\ref{sec:routing-data}, the
curriculum progressively introduces higher-scored examples while retaining
lower-scored examples in later stages.

The model actually served, however, also reflects user overrides, service
availability, and deployment policies. Its identity alone is therefore an
imperfect proxy for the capability an interaction requires. We instead
re-estimate capability demand from the request and relevant interaction
history available before the first supervised assistant response. We assign
this estimate to the complete example $x_i$ defined in
Section~\ref{sec:method-agentic-supervision}, using it as a turn-level ordering
proxy rather than a difficulty label for each individual action. This changes
when the example is presented, without changing its recorded assistant targets.

\paragraph{Routing scores.}
For each example, the router provides a tier assignment and a normalized
vector of scores expressing relative support for C0--C3.
Let $k_i\in\{0,1,2,3\}$ be the assigned tier index and $\pi_{i,k}$ the
normalized score for tier C$k$, with $\pi_{i,k}\geq 0$ and
$\sum_{k=0}^{3}\pi_{i,k}=1$. We use the assigned tier index for hard
ordering and the score-weighted mean tier index for soft ordering:
\begin{equation}
  s_i =
  \begin{cases}
    k_i, & \text{hard ordering}, \\
    \displaystyle\sum_{k=0}^{3} k\,\pi_{i,k}, & \text{soft ordering}.
  \end{cases}
  \label{eq:curriculum-ordering-score}
\end{equation}
The soft score can distinguish examples with the same assigned tier by
incorporating support for the other tiers. We use $s_i$ to construct the
curriculum described next, not to reweight the SFT loss.

\begin{figure}[t]
  \centering
  \includegraphics[width=\linewidth]{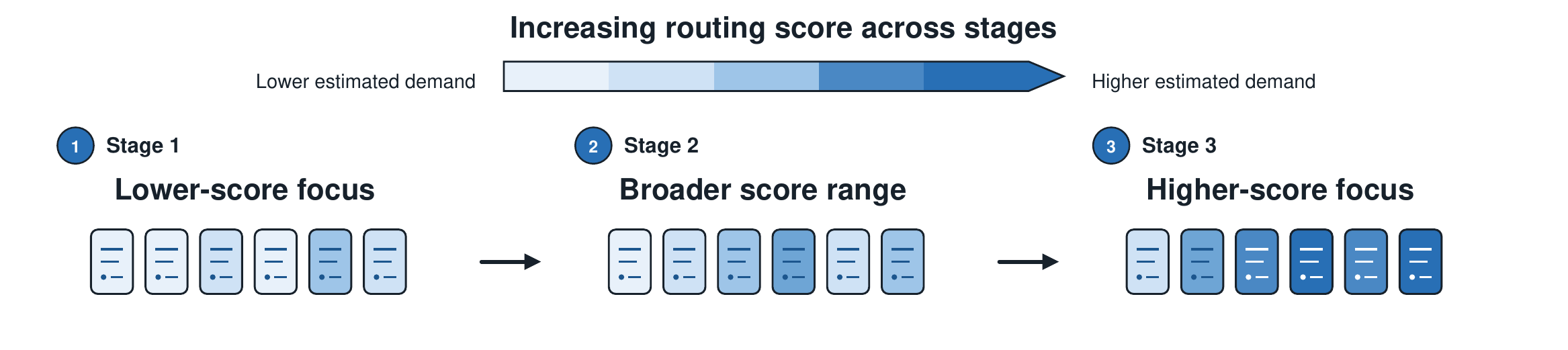}
  \caption{\textbf{Routing-guided curriculum.} Routing scores serve as a proxy for
  capability demand. The training mix shifts toward higher-scored examples
  over three approximately equal-sized stages, while some lower-scored
  examples are reserved for later stages. Color intensity schematically
  indicates the routing-score distribution within each stage.}
  \label{fig:routing-curriculum}
\end{figure}

\paragraph{Curriculum scheduling.}
We train over three stages, gradually introducing examples with higher
routing scores. Each stage contains roughly one third of the examples,
with some lower-scored examples reserved for later stages. Reserving
lower-scored examples for later stages prevents the end of training from
being dominated exclusively by high-demand interactions. Every example
is used once per pass. Figure~\ref{fig:routing-curriculum} illustrates the
schedule. Training follows the same masked SFT objective throughout,
without resetting the optimizer or restarting the learning-rate schedule
between stages.

\subsection{Routing-Guided On-Policy Distillation}
\label{sec:method-opd}

The SFT objective in Section~\ref{sec:method-agentic-supervision} learns
from recorded assistant responses, whereas a deployed student conditions
on its own generated prefixes. On-policy distillation (OPD)
\citep{agarwal2024onpolicy} provides teacher supervision on these
student-generated prefixes.

The starting contexts provide the learning material for distillation:
they determine which tasks and interaction states the student encounters.
We extend the routing-guided curriculum in
Section~\ref{sec:method-routing-curriculum} to organize this material by
routing-estimated capability demand. Routing controls the progression of
learning material, while on-policy supervision follows the student's
evolving behavior.

\paragraph{Routing-guided context scheduling.}
We use recorded contexts immediately before assistant responses as
generation starting points. Each starting state is scored from the request
and interaction history available at that point, following
Equation~\ref{eq:curriculum-ordering-score}. We apply the same three-stage
allocation as in SFT, progressively introducing higher-scored contexts
while reserving some lower-scored contexts for later stages.
Let $\rho_j$ denote the distribution over context batches induced by
this allocation at stage $j\in\{1,2,3\}$.

\paragraph{Student generation and teacher supervision.}
At stage $j$, a student checkpoint $p_{\bar\theta}$ generates one assistant
response for each context in a batch $\mathcal{C}$ drawn from $\rho_j$.
The resulting response batch $\mathcal{R}$ may contain reasoning, tool
calls, or visible text. A fixed teacher supplies a next-token distribution
at each position, conditioned on the corresponding context and the
student's preceding response tokens. Both models render the same recorded
messages and tools using their respective native templates, with response
token IDs aligned for scoring. We refresh the rollout checkpoint as
training proceeds, so later contexts receive teacher supervision on more
recent student behavior. The rollout parameters $\bar\theta$ remain fixed
while optimizing the student on the collected responses.

\begin{figure}[t]
  \centering
  \includegraphics[width=\linewidth]{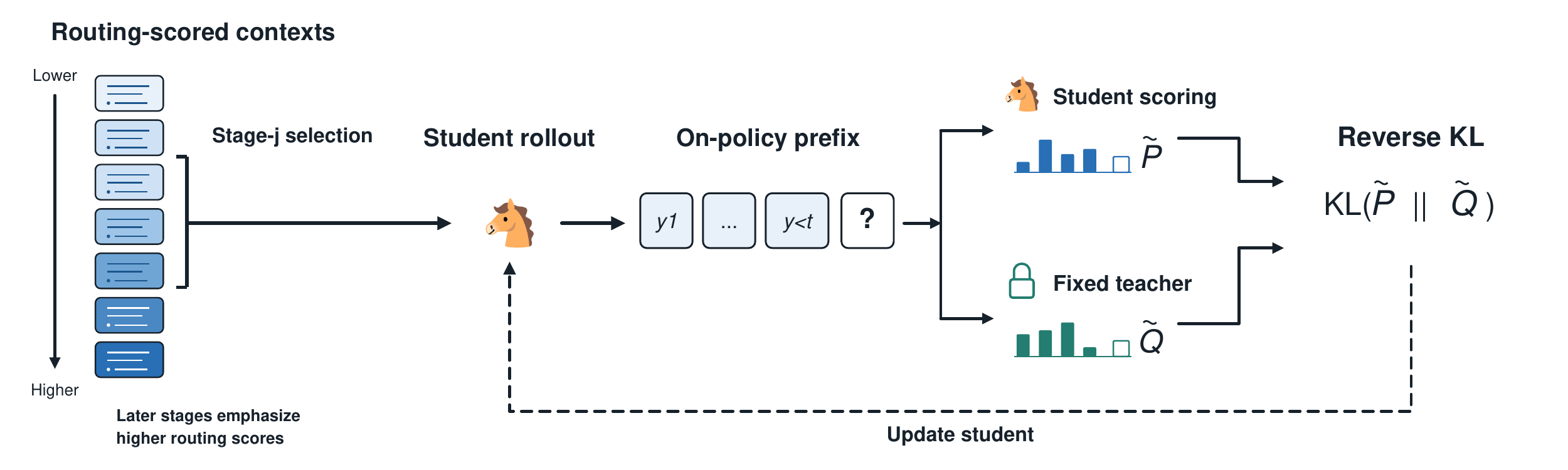}
  \caption{\textbf{Routing-guided on-policy distillation.}
  Routing scores schedule the recorded starting contexts over three stages,
  with lower-scored contexts reserved for later stages. The student generates
  responses from these contexts. At each generated prefix, the student and
  a fixed teacher provide next-token distributions for the reverse-KL
  objective in Equation~\ref{eq:opd-response-objective}.
  Both distributions use the same top-$K$ candidate tokens and a bin for
  the remaining probability mass. Only the student is updated; refreshed
  student checkpoints generate subsequent responses. Rollout and scoring
  use the same student model, with parameters updated during training.
  Token strips and probability bars are schematic.}
  \label{fig:routing-opd}
\end{figure}

\paragraph{Distillation objective.}
Let $P_{\theta,r,t}$ and $Q_{r,t}$ denote the current student and fixed
teacher distributions at position $t$ of generated response $r$.
For compact supervision, we retain the rollout student's top-$K$ candidate
tokens at each position and aggregate all remaining probability mass into
one additional bin. Both models use the same candidate set and
full-vocabulary probabilities, yielding coarsened distributions
$\widetilde{P}_{\theta,r,t}$ and $\widetilde{Q}_{r,t}$ over $K+1$ bins.
For a batch $\mathcal{R}$, we minimize the response-normalized reverse KL:
\begin{equation}
  \mathcal{L}_{\mathrm{OPD}}(\theta;\mathcal{R})
  = \frac{1}{\sum_{r\in\mathcal{R}}w_r}
    \sum_{r\in\mathcal{R}}\frac{w_r}{L_r}
    \sum_{t=1}^{L_r}
      D_{\mathrm{KL}}\!\left(
        \widetilde{P}_{\theta,r,t}\,\Vert\,\widetilde{Q}_{r,t}\right).
  \label{eq:opd-response-objective}
\end{equation}
Here $L_r$ is the retained response length and $w_r$ is a fixed response
weight, equal to one in the unweighted setting. Each response contributes
its average token-level divergence; prompt tokens and padding receive no
loss. Gradients are taken through the current student's probabilities on
the collected prefixes, holding the generated tokens, candidate IDs, and
teacher scores fixed.

The stage-wise objective combines the context distribution $\rho_j$ with
student generation and the response-level loss defined above:
\begin{equation}
  \mathcal{L}_{\mathrm{R\text{-}OPD}}^{(j)}(\theta)
  =
  \mathbb{E}_{\substack{
    \mathcal{C}\sim\rho_j\\
    \mathcal{R}\sim p_{\bar\theta}(\cdot\mid\mathcal{C})
  }}
  \left[\mathcal{L}_{\mathrm{OPD}}(\theta;\mathcal{R})\right],
  \qquad j\in\{1,2,3\}.
  \label{eq:routing-opd-stage-objective}
\end{equation}
Figure~\ref{fig:routing-opd} summarizes the complete training process.

\section{Results and Analysis}
\label{sec:results-and-analysis}

This section analyzes \modelname{} from three complementary perspectives:
benchmark performance, agent execution behavior, and training data.
We first report the main results of the \modelname{}-4B and \modelname{}-9B models across agentic,
coding, and instruction-following benchmarks, examining whether training
provides consistent gains at a fixed model scale and whether the 9B variant
provides additional improvements over the 4B variant. We then study
representative agent trajectories to characterize qualitative differences
in evidence acquisition, constraint tracking, execution verification,
failure recovery, and strategy adaptation. Finally, we study the effects
of trajectory source and supervision scale to characterize how the
composition and quantity of agent interaction data relate to downstream
performance.


\paragraph{Benchmarks.}

We evaluate \modelname{} on a diverse suite of benchmarks,
organized into three top-level categories aligned with the evaluation
tables: agentic capabilities, code generation, and instruction following.
The agentic category further covers two complementary aspects:
end-to-end agent execution and tool use with interactive task completion.

\begin{itemize}

\item \textbf{Agentic:}
For end-to-end agent execution,
QwenClawBench~\citep{qwenclawbench2026} targets realistic
OpenClaw tasks, WorkBuddy Bench~\citep{workbuddybench2026}
covers multi-domain workplace scenarios, PinchBench~
\citep{pinchbench2026} focuses on standardized OpenClaw
workflows, and VitaBench~\citep{vitabench2025} examines
multi-turn interactions in daily-life service scenarios.
For tool use and interactive task completion,
BFCL V4~\citep{patil2025bfcl} measures function-calling
and agentic tool-use capabilities, whereas
$\tau^2$-Bench~\citep{barres2025tau2} evaluates multi-turn
task completion involving user--agent--tool interactions
in the Airline, Retail, and Telecom domains.

\item \textbf{Coding:}
HumanEval~\citep{chen2021evaluating} measures the functional
correctness of programs synthesized from natural-language
specifications, while LiveCodeBench v6~
\citep{jain2024livecodebench} uses recent
competition-style programming problems to assess coding
performance.

\item \textbf{Instruction following:}
IFEval~\citep{zhou2023instruction} tests compliance with
explicitly verifiable instructions, whereas IFBench~
\citep{pyatkin2025generalizing} focuses on generalization
to diverse and previously unseen constraints.

\end{itemize}


\paragraph{Baselines.}
\label{sec:exp-baselines}

We evaluate \modelname{} at two model scales against a representative
set of strong and competitive open-weight models. The selected baselines
include both established general-purpose models and models specifically
optimized for agentic capabilities. Our evaluation focuses exclusively
on text-based tasks, including reasoning, instruction following, tool
use, coding, and multi-step agent interaction. For models that support
multiple modalities, only their text interfaces are evaluated.

For the 4B track, we compare \modelname{}-4B with
\textbf{Qwen3.5-4B}~\citep{qwen3.5},
\textbf{Spark-X2.5-4B}~\citep{sparkx2.5},
\textbf{Gemma-4-E4B-it}~\citep{gemmateam2026gemma4technicalreport},
\textbf{Nanbeige-4.2-3B}~\citep{lab2026nanbeige423bunlockingagenticcapabilities}, and
\textbf{Agents-A1-4B}~\citep{bai2026scalinghorizonparametersreaching}.

For the 9B track, we compare \modelname{}-9B with
\textbf{Granite-4.2-8B}~\citep{granite2026},
\textbf{Qwen3.5-9B}~\citep{qwen3.5},
\textbf{Ornith-1.5-9B}~\citep{ornith_1_5},
\textbf{Gemma-4-12B-it}~\citep{gemmateam2026gemma4technicalreport}, and
\textbf{Muse-Glimmer-30B}~\citep{museglimmer30b}
as contextual reference models.

Unless otherwise noted, within each benchmark, all models evaluated
by us use the same benchmark-specific harness, tool interfaces,
context limits, and interaction budgets. We use the official chat
template of each model. Results taken from an official blog post or
technical report are marked with $^{*}$ in the corresponding table
and are included as reference results rather than as measurements
produced under our evaluation pipeline.

\paragraph{Evaluation Configurations}

For each model evaluated by us, we use the inference parameters
recommended in its official documentation. For \modelname{} and its
Qwen3.5-4B and Qwen3.5-9B base models, we follow the Qwen3.5
recommendations: temperature $=1.0$, top-$p=0.95$, top-$k=20$,
min-$p=0.0$, presence penalty $=1.5$, and repetition penalty $=1.0$.
All variants of \modelname{} are deployed using SGLang v0.5.17.
Both \modelname{} and these Qwen3.5 baselines are evaluated in thinking mode, with
\texttt{chat\_template\_kwargs} configured as follows:
\texttt{\{"enable\_thinking": true,
"force\_nonempty\_content": true\}}.
The maximum output length is set to 51200 tokens for IFEval, IFBench,
HumanEval, and LiveCodeBench v6, and to 32768 tokens for all other benchmarks.

\textbf{Harness-based Agents.}
QwenClawBench and PinchBench are evaluated using OpenSquilla~\cite{opensquilla2026} as the
agent harness. WorkBuddy Bench is evaluated with its official native
harness, while VitaBench is evaluated using its official framework.
For VitaBench, we use DeepSeek-V4-Flash as both the user-simulator
model and the judge model because the originally recommended models
are no longer available.

\textbf{Tool Use and Interactive Agents.}
BFCL V4 is evaluated using the native function-calling mode.
For $\tau^2$-Bench, we use its official evaluation framework across
the Airline, Retail and Telecom.

\textbf{Repeated Runs and Aggregation.}
We perform three independent runs for QwenClawBench, WorkBuddy Bench,
and $\tau^2$-Bench, and report the arithmetic mean across runs.
PinchBench and VitaBench are each evaluated with a single run.
For the remaining benchmarks, we follow their official evaluation
and scoring protocols.

\paragraph{Overall Results.}

Tables~\ref{tab:results-4b} and~\ref{tab:results-large} summarize
the performance of the 4B and 9B models across agentic,
coding, and instruction-following benchmarks. Two main patterns emerge.
First, training provides broad improvements at both model scales.
Second, increasing model scale yields further gains, with the additional
benefits concentrated primarily on tasks involving multi-step interaction,
execution feedback, and challenging code generation.

\begin{table*}[t]
  \centering
  \small
  \setlength{\tabcolsep}{1.8pt}
  \renewcommand{\arraystretch}{1.12}

  \caption{
    Comparison of \modelname{}-4B with representative 4B-scale
    open-weight models across agentic, coding, and
    instruction-following benchmarks.
    Higher is better.
    The best and second-best available results in each column
    are shown in bold and underlined, respectively.
  }
  \label{tab:results-4b}

  \begin{adjustbox}{max width=\textwidth}
  \begin{tabular}{@{}l*{11}{c}@{}}
    \toprule

    \multirow{2}{*}{\textbf{Model}}
      & \multicolumn{6}{c}{\textbf{Agentic}}
      & \multicolumn{2}{c}{\textbf{Coding}}
      & \multicolumn{2}{c}{\textbf{Instruction Following}}
      & \multirow{2}{*}{\textbf{Avg.}} \\

    \cmidrule(lr){2-7}
    \cmidrule(lr){8-9}
    \cmidrule(lr){10-11}

      & \shortstack{
          \footnotesize\textbf{BFCL}\\[-1pt]
          \footnotesize\textbf{v4}
        }
      & \shortstack{
          \footnotesize\textbf{Vita}\\[-1pt]
          \footnotesize\textbf{Bench}
        }
      & \shortstack{
          \footnotesize$\boldsymbol{\tau^2}$\\[-1pt]
          \footnotesize\textbf{Bench}
        }
      & \shortstack{
          \footnotesize\textbf{Pinch}\\[-1pt]
          \footnotesize\textbf{Bench}
        }
      & \shortstack{
          \footnotesize\textbf{WorkBuddy}\\[-1pt]
          \footnotesize\textbf{Bench}
        }
      & \shortstack{
          \footnotesize\textbf{QwenClaw}\\[-1pt]
          \footnotesize\textbf{Bench}
        }
      & \shortstack{
          \footnotesize\textbf{Human}\\[-1pt]
          \footnotesize\textbf{Eval}
        }
      & \shortstack{
          \footnotesize\textbf{LiveCode}\\[-1pt]
          \footnotesize\textbf{Bench v6}
        }
      & \shortstack{
          \footnotesize\textbf{IF}\\[-1pt]
          \footnotesize\textbf{Bench}
        }
      & \shortstack{
          \footnotesize\textbf{IF}\\[-1pt]
          \footnotesize\textbf{Eval}
        }
      & \\

    \midrule

    Qwen3.5-4B
      & 61.02
      & 21.50
      & 84.29
      & 71.19
      & 24.62
      & 38.47
      & 87.20
      & 53.71
      & 60.33
      & 87.06
      & 58.94 \\

    Spark-X2.5-4B
      & \underline{63.71}
      & \underline{37.00}
      & 77.72
      & 62.37
      & 26.47
      & \underline{43.52}
      & 92.07
      & 54.86
      & \textbf{73.33}
      & \textbf{91.13}
      & 62.22 \\

    Gemma-4-E4B-it
      & 47.18
      & 5.00
      & 43.60
      & 47.60
      & 11.65
      & 22.98
      & 84.76
      & 52.00
      & 40.00
      & 74.68
      & 42.95 \\

    Nanbeige-4.2-3B
      & \textbf{67.28}
      & 31.50
      & \underline{85.08}
      & 66.78
      & 21.03
      & 40.66
      & \textbf{98.78}
      & \textbf{72.50$^{*}$}
      & 55.00
      & 84.47
      & \underline{62.31} \\

    Agents-A1-4B
      & 46.60
      & \textbf{39.25}
      & 81.00
      & \underline{75.07}
      & \underline{33.37}
      & 43.16
      & 92.68
      & 56.57
      & 63.33
      & 83.55
      & 61.46 \\

    \midrule

    \rowcolor[gray]{0.93}
    \textbf{\modelname{}-4B}
      & 61.79
      & 32.00
      & \textbf{88.46}
      & \textbf{77.33}
      & \textbf{34.41}
      & \textbf{44.68}
      & \underline{96.95}
      & \underline{59.43}
      & \underline{65.33}
      & \underline{88.35}
      & \textbf{64.87} \\

    \bottomrule
  \end{tabular}
  \end{adjustbox}

  \vspace{3pt}
  \begin{minipage}{0.99\textwidth}
    \footnotesize
    \textit{Notes.}
    $^{*}$ denotes a result reported in the corresponding model's
    official blog post or technical report.
  \end{minipage}
\end{table*}

\begin{table*}[t]
  \centering
  \small
  \setlength{\tabcolsep}{1.8pt}
  \renewcommand{\arraystretch}{1.12}

  \caption{
    Comparison of \modelname{}-9B with representative larger-scale
    models across agentic, coding, and instruction-following benchmarks.
    Higher is better.
    The best and second-best available results in each column
    are shown in bold and underlined, respectively.
  }
  \label{tab:results-large}

  \begin{adjustbox}{max width=\textwidth}
  \begin{tabular}{@{}l*{11}{c}@{}}
    \toprule

    \multirow{2}{*}{\textbf{Model}}
      & \multicolumn{6}{c}{\textbf{Agentic}}
      & \multicolumn{2}{c}{\textbf{Coding}}
      & \multicolumn{2}{c}{\textbf{Instruction Following}}
      & \multirow{2}{*}{\textbf{Avg.}} \\

    \cmidrule(lr){2-7}
    \cmidrule(lr){8-9}
    \cmidrule(lr){10-11}

      & \shortstack{
          \footnotesize\textbf{BFCL}\\[-1pt]
          \footnotesize\textbf{v4}
        }
      & \shortstack{
          \footnotesize\textbf{Vita}\\[-1pt]
          \footnotesize\textbf{Bench}
        }
      & \shortstack{
          \footnotesize$\boldsymbol{\tau^2}$\\[-1pt]
          \footnotesize\textbf{Bench}
        }
      & \shortstack{
          \footnotesize\textbf{Pinch}\\[-1pt]
          \footnotesize\textbf{Bench}
        }
      & \shortstack{
          \footnotesize\textbf{WorkBuddy}\\[-1pt]
          \footnotesize\textbf{Bench}
        }
      & \shortstack{
          \footnotesize\textbf{QwenClaw}\\[-1pt]
          \footnotesize\textbf{Bench}
        }
      & \shortstack{
          \footnotesize\textbf{Human}\\[-1pt]
          \footnotesize\textbf{Eval}
        }
      & \shortstack{
          \footnotesize\textbf{LiveCode}\\[-1pt]
          \footnotesize\textbf{Bench v6}
        }
      & \shortstack{
          \footnotesize\textbf{IF}\\[-1pt]
          \footnotesize\textbf{Bench}
        }
      & \shortstack{
          \footnotesize\textbf{IF}\\[-1pt]
          \footnotesize\textbf{Eval}
        }
      & \\

    \midrule

    Qwen3.5-9B
      & 64.88
      & 31.25
      & \underline{88.04}
      & \underline{74.55}
      & 39.60
      & 44.04
      & 92.68
      & 65.14
      & 66.33
      & 89.46
      & 65.60 \\

    Granite-4.2-8B
      & 52.06
      & 23.00
      & 62.28
      & 56.93
      & 35.07
      & 37.01
      & 96.34
      & \underline{72.00}
      & \underline{78.00}
      & 92.98
      & 60.57 \\

    Ornith-1.5-9B
      & \underline{65.03}
      & 26.75
      & 83.68
      & 68.22
      & 29.29
      & \underline{47.27}
      & 93.90
      & 47.43
      & 40.00
      & 71.35
      & 57.29 \\

    Gemma-4-12B-it
      & 62.06
      & 36.50
      & 59.37
      & 58.89
      & 29.65
      & 43.53
      & \textbf{100.00}
      & \textbf{73.14}
      & 77.67
      & \textbf{94.27}
      & 63.51 \\

    Muse-Glimmer-30B
      & 53.74
      & \textbf{48.50}
      & 76.64
      & 71.35
      & \textbf{45.85}
      & 46.11
      & \underline{98.17}
      & 65.71
      & \textbf{78.67}
      & \underline{93.90}
      & \underline{67.86} \\

    \midrule

    \rowcolor[gray]{0.93}
    \textbf{\modelname{}-9B}
      & \textbf{67.43}
      & \underline{42.25}
      & \textbf{90.82}
      & \textbf{82.25}
      & \underline{40.15}
      & \textbf{48.73}
      & \underline{98.17}
      & 65.14
      & 66.33
      & 89.09
      & \textbf{69.04} \\

    \bottomrule
  \end{tabular}
  \end{adjustbox}

\end{table*}

\textbf{Broad gains at both model scales.}
At the 4B scale, \modelname{}-4B outperforms Qwen3.5-4B on every benchmark
for which both models have available results. The gains are particularly
pronounced on harness-based agent tasks and coding benchmarks, while
consistent improvements are also observed in tool use and instruction
following. This pattern indicates that the improvement is not driven by
a small subset of metrics, but extends across diverse task formats and
capability dimensions.

At the 9B scale, \modelname{}-9B improves over Qwen3.5-9B on most of the
available benchmarks. The gains are again concentrated on harness-based
agent tasks, tool interaction, and selected coding tasks. By contrast,
performance on instruction-following benchmarks remains largely stable,
with one metric showing a minor decrease. These results indicate that
training remains effective for the stronger 9B base model, although its
marginal benefits are concentrated more heavily on interactive execution
than on relatively static instruction compliance.

\textbf{Larger configurations remain beneficial.}
Across all benchmarks with available results for both variants,
\modelname{}-9B consistently outperforms \modelname{}-4B. The additional
gains, however, are not uniformly distributed across capabilities.
They are more pronounced on harness-based agent tasks, function calling,
and challenging coding benchmarks, while the differences on
instruction-following evaluations are comparatively small.

This pattern suggests that increasing model scale remains particularly
valuable when successful task completion requires sustained state
tracking, interaction with external tools, and revision of intermediate
decisions in response to execution feedback. By comparison, the gap
between the two model scales is substantially smaller on tasks that
primarily assess compliance with explicit, relatively static
instructions.

Notably, \modelname{}-4B already matches or exceeds Qwen3.5-9B on several
benchmarks, suggesting that training can compensate for part of the
performance gap associated with model scale.

The benchmark-level pattern is also reflected in the agent trajectories:
the larger model is more effective at maintaining iterative verification
loops, recovering from failed actions, and adapting its strategy in
response to environmental feedback.

\subsection{Agent Trace Analysis}
\label{sec:agent-level-analysis}

The main results reveal two related patterns: training improves
performance at both model scales, while the larger model retains an
advantage on interaction- and execution-intensive tasks. To understand
the behavioral mechanisms underlying these differences, we examine
representative agent trajectories.

\paragraph{Training closes the end-to-end execution loop.}

The same-scale gains are reflected not only in final benchmark scores,
but also in whether the model can organize a sequence of locally
plausible actions into a complete and coherent workflow.

In a representative QwenClawBench project-scheduling trajectory,
Qwen3.5-4B identifies the relevant files in the working directory but
fails to inspect a manager's email containing an updated dependency
constraint. As a result, it plans from outdated information, produces
an invalid schedule, and writes the resulting artifact to an unintended
location.

By contrast, \modelname{}-4B retrieves the additional evidence,
recognizes the updated dependency, recomputes and verifies the schedule,
and saves the final artifact to the required path.

\paragraph{Scale pays off under feedback and failure.}

Although training substantially strengthens the 4B model, the larger
model remains more robust on tasks that require iterative debugging,
recovery from execution failures, and maintenance of state over longer
action sequences.

In a WorkBuddy code-repair trajectory, \modelname{}-4B stops after a
single implementation attempt without establishing an effective testing
and repair loop, leaving an error in its handling of thread-execution
semantics. \modelname{}-9B instead follows a complete
edit--test--inspect--repair cycle and repeatedly incorporates execution
feedback until the verifier passes.

The difference is not limited to the quality of the initial
implementation. It also concerns whether the model treats test outcomes
as inputs to subsequent decisions and converts a failed attempt into
evidence for further repair. Rather than producing a one-shot solution,
the larger model is better able to organize environmental feedback into
a continuing problem-solving process.

A PinchBench data-analysis trajectory reveals a complementary advantage
in strategy adaptation under environmental constraints. After
discovering that \texttt{pandas} is unavailable, \modelname{}-4B
repeatedly attempts dependency installation, manual CSV parsing, and
local script patching. These attempts fail to resolve the underlying
constraint, introduce additional errors, and ultimately prevent the
model from producing the requested report.

\modelname{}-9B instead recognizes that the original approach is
blocked, abandons repeated attempts to use the unavailable dependency,
and switches to Python's standard \texttt{csv} and mathematical
libraries. It then completes both the analysis and the final report.
Relative to the 4B trajectory, the 9B trajectory reduces the number
of model requests, execution time, and token usage by approximately
$70.8\%$, $76.7\%$, and $83.6\%$, respectively.

These cases suggest that the larger model's advantage does not arise
from executing more actions or conducting a broader but undirected
search. Instead, it is better able to identify unproductive trajectories,
revise its strategy in response to environmental feedback, and allocate
a limited interaction budget to actions that directly contribute to
task completion. The WorkBuddy case highlights iterative verification
and error repair, whereas the PinchBench case highlights strategy
recovery under environmental constraints. Together, they show that the
larger model is more effective at sustaining closed-loop execution under
feedback and failure.
\subsection{Data Analysis}
\label{sec:data-analysis}

We analyze agentic supervision along two complementary axes: whether
routing-harness trajectories transfer more effectively than public agent data
under matched training conditions, and how performance changes as the amount
of routing-harness supervision increases. Unless otherwise specified, all
controlled experiments in this section use the 4B model initialized from
Qwen3.5-4B. We use five benchmarks that are available for every checkpoint
and span coding, instruction following, function calling, and multi-turn tool use.

\paragraph{Routing-harness data versus public agent data.}

To examine whether the source of agentic trajectories matters beyond the
curriculum algorithm itself, we compare our routing-harness data with Toucan,
a public synthetic tool-agent dataset~\citep{xu2025toucan}. Before curriculum
construction, examples from both sources receive capability-demand scores from
the same offline routing-labeling procedure. Both runs start from the same
Qwen3.5-4B checkpoint and use the same curriculum schedule, optimizer settings,
random seed, packing method, and evaluation protocol, with closely matched
training budgets.

\begin{table*}[t]
  \centering
  \small
  \setlength{\tabcolsep}{3.2pt}
  \renewcommand{\arraystretch}{1.12}
  \caption{Comparison of routing-harness trajectories with public tool-agent
  data under the same routing-guided training configuration. Higher is better;
  the final row reports absolute percentage-point differences, and Avg. is the
  unweighted mean across the five benchmarks.}
  \label{tab:routing-data-vs-public}
  \begin{adjustbox}{max width=\textwidth}
  \begin{tabular}{@{}lrrrrrr@{}}
    \toprule
    \textbf{Training data} & \textbf{LCB} & \textbf{HE} & \textbf{IF} &
    \textbf{BFCL} & \textbf{$\tau^2$} & \textbf{Avg.} \\
    \midrule
    Public agent data (Toucan) & 49.14 & 87.80 & 56.33 & 54.77 & 73.54 & 64.32 \\
    \textbf{Routing-harness data} & \textbf{53.14} & \textbf{96.34} &
    \textbf{61.33} & \textbf{57.20} & \textbf{84.85} & \textbf{70.57} \\
    \midrule
    \textbf{Difference (ours $-$ public)} & \textbf{+4.00} & \textbf{+8.54} &
    \textbf{+5.00} & \textbf{+2.43} & \textbf{+11.31} & \textbf{+6.26} \\
    \bottomrule
  \end{tabular}
  \end{adjustbox}
\end{table*}

The routing-harness checkpoint is stronger on all five comparable benchmarks,
improving their unweighted average by 6.26 points.
The largest gains appear on HumanEval ($+8.54$) and $\tau^2$-Bench
($+11.31$), while function calling, instruction following, and coding also
improve. These results show that
routing-mediated interactions provide stronger and more transferable agentic
supervision than public synthetic trajectories under the same routing-guided
training recipe. The sources differ in sequence composition, while their
overall training budgets remain comparable.

\paragraph{Scaling routing-harness supervision.}

\begin{wrapfigure}{r}{0.46\textwidth}
\vspace{-4mm}
  \centering
  \includegraphics[width=\linewidth]{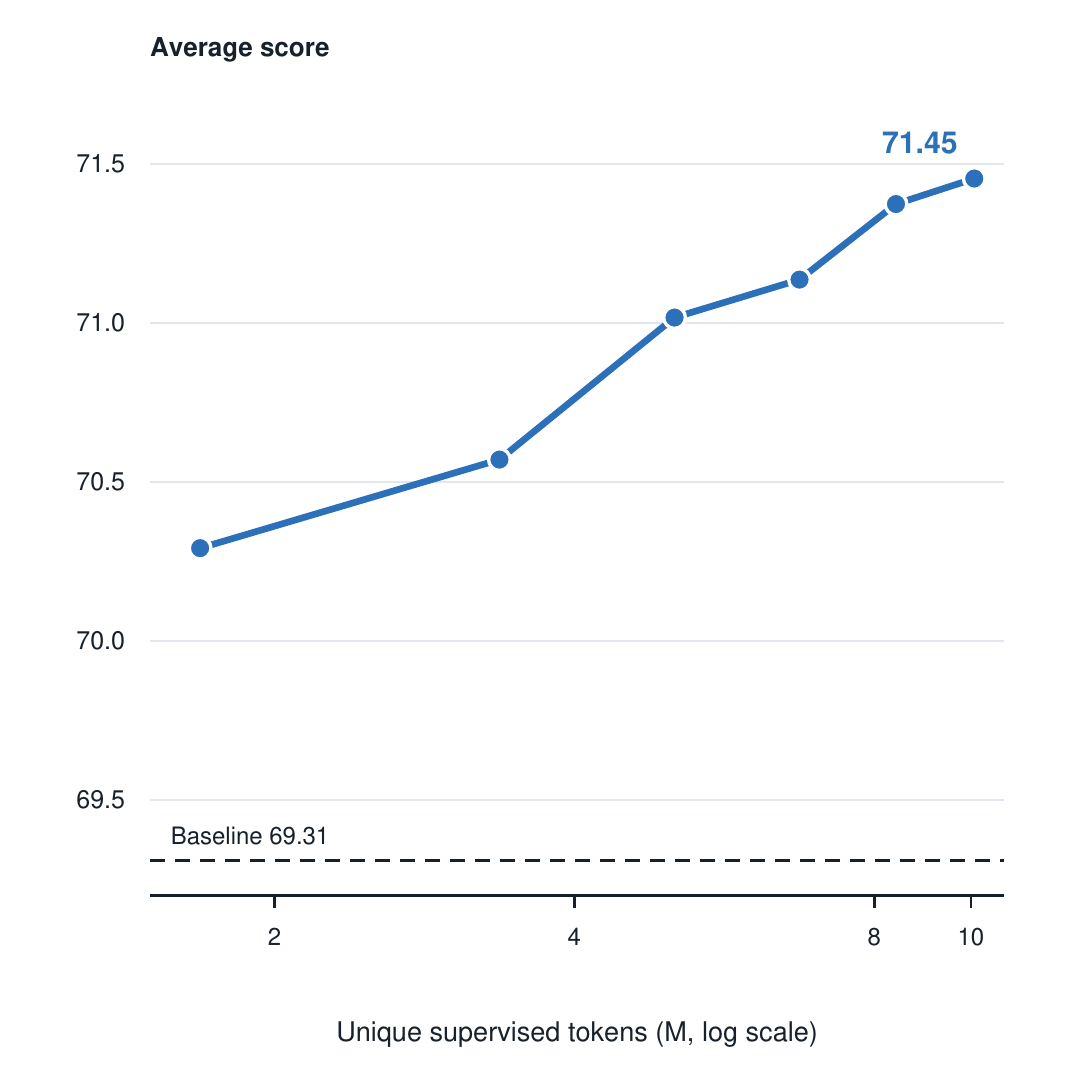}
  \caption{\textbf{Scaling routing-harness supervision.}
  }
  \label{fig:routing-data-scaling}
  \vspace{-8mm}
\end{wrapfigure}

We next examine how performance changes as the amount of routing-harness
supervision increases. Starting from a single quality-ranked trajectory pool,
we construct strictly nested subsets: every larger subset contains all
examples from the preceding subset. Model initialization, optimization,
packing, and the number of passes over each subset are held fixed. This setup
isolates the effect of adding unique supervision while preserving the data
selection policy.

Figure~\ref{fig:routing-data-scaling} reports the unweighted average across five
benchmarks available for every checkpoint: LiveCodeBench, HumanEval, IFBench,
BFCL V4, and $\tau^2$-Bench. The development-suite average increases steadily
from 69.31 for the base model to 71.45 at the largest data scale shown. Data
scale is measured in unique supervised tokens and shown on a logarithmic axis.

These results indicate that scaling high-quality agentic supervision can
produce consistent aggregate gains over the evaluated range. They also
support treating data quantity as a capability-dependent allocation decision:
the useful operating point is determined jointly by supervision quality,
coverage, and the capability profile targeted by post-training.

\section{Task-Specific Adaptation: NeoHorse-Jev}
\label{sec:neohorse-jev}

Agentic post-training provides a foundation not only for generating responses
and tool calls, but also for specialized components within an agent workflow.
We develop \textbf{NeoHorse-Jev-4B} by adapting NeoHorse-1-4B into a structured
decision model. Given application state, a question, and a set of admissible
answers, it returns a decision and an associated probability distribution.
This interface supports request routing, tool selection, condition checking,
and outcome assessment without generating a free-form response.

\paragraph{Decision interface and adaptation.}
NeoHorse-Jev uses prefill-only inference: the backbone processes the supplied
state and question, and a learned decision readout scores the available
answers, building on the open-source decision readout in
Kev~\cite{palmer2026kev}. Following the typed-question interface of
Jev~\cite{typesafe2026primitives}, it supports three answer types:
\emph{Choice} selects from named
alternatives; \emph{Noul} answers a yes/no question; and \emph{Score} predicts
over application-defined ordered rating levels. Probabilities remain available
to the calling application, allowing software to apply thresholds, choose a
fallback, or request additional evidence rather than treating every prediction
as an instruction to execute.

Training adapts the agentic backbone and the decision readout using supervised
decision examples. Each example pairs a state with one or more questions,
candidate answers, and reference decisions. The supervision targets these
structured decisions rather than an autoregressive explanation. This changes
the output interface while retaining NeoHorse's language representation as
the foundation for interpreting requests, context, and alternatives. The
result is a reusable decision component; the surrounding harness retains
responsibility for executing actions and updating state.

\paragraph{Text decision evaluation.}
Table~\ref{tab:neohorse-jev} reports six evaluation groups.
JevBench~\cite{standhartinger2026jevbench} uses its public family-macro score;
Kev~\cite{palmer2026kev} averages six development/test suite scores.
OpenJev text denotes the static text evaluation suite released with
AlexWortega's OpenJev~\cite{alexwortega2026openjev}; we average its 19 task
scores. Nimble~\cite{nimble2026}, VitaminC~\cite{schuster2021vitaminc}, and
MASSIVE~\cite{fitzgerald2022massive} use frozen, adapted subsets of 282, 599,
and 350 examples, respectively.
The six-group aggregate weights groups equally. Missing results are not
imputed, and the original NeoHorse-1-4B is compared on the three common subsets
using answer-letter logits with thinking disabled, whereas NeoHorse-Jev uses
its decision readout.

\begin{table}[t]
\centering
\caption{\textbf{Structured text decisions.} Scores on a 0--100 scale;
higher is better. AVG equally weights the six evaluation groups and is
reported only for complete rows. The three rightmost benchmarks are frozen
evaluation subsets, not their official full datasets.}
\label{tab:neohorse-jev}
\small
\setlength{\tabcolsep}{4pt}
\begin{adjustbox}{max width=\textwidth}
\begin{tabular}{@{}lrrrrrrr@{}}
\toprule
Model & JevBench & Kev & \shortstack{OpenJev\\text} & Nimble & VitaminC & MASSIVE & AVG \\
\midrule
Open-Jev-9B~\cite{zefancai2026openjev9b} & 77.13 & 77.87 & 65.39 & 80.50 & 68.28 & 84.86 & 75.67 \\
Kev-4B~\cite{palmer2026kev4b} & 73.71 & 81.47 & 54.75 & 73.40 & 76.46 & 85.71 & 74.25 \\
Laya English~\cite{convai2026laya} & 55.82 & 61.30 & 40.07 & 45.04 & 78.63 & 68.57 & 58.24 \\
NeoHorse-1-4B & -- & -- & -- & 69.15 & 63.27 & 82.86 & -- \\
\textbf{NeoHorse-Jev-4B} & 75.73 & 81.92 & 58.74 & 87.23 & 77.13 & 85.43 & 77.70 \\
\bottomrule
\end{tabular}
\end{adjustbox}
\end{table}

NeoHorse-Jev achieves a six-group aggregate of 77.70. On the three common
subsets, adaptation raises the equal-weight mean accuracy from 71.76\% to
83.26\%, a gain of 11.50 percentage points. The improvement covers all three
subsets, demonstrating the usefulness of adapting an agentic language model
to an application-defined decision interface. The comparison evaluates the
complete adapted system, including its readout, rather than isolating one
training objective.

The model and decision-aware runtime are available at
\url{https://huggingface.co/TokenRhythm/NeoHorse-Jev-4B}.

\section{Conclusion and Discussion}\label{sec:conclusion}
Recursive self-improvement requires a concrete mechanism through which a
system observes its own capabilities and turns that evidence into the next
round of learning. This report presented \modelname{}, a family of
agent-native models built on the observation that a deployed routing
harness already contains such a mechanism. Beyond serving user requests,
the harness produces three reusable signals: execution trajectories that
ground training in real interaction, routing signals that characterize
capability demand, and recorded outcomes that
reveal where the model still falls short. Our system realizes this idea
through three connected components. On the data side, harness interactions
are converted into user-turn training examples that preserve interleaved
reasoning, tool calls, and harness context, and are admitted through
structural validation, six-dimensional semantic evaluation, and
subscene-level labeling. On the method side, routing scores organize this
experience into a three-stage curriculum for supervised fine-tuning and
extend to routing-guided on-policy distillation, in which a teacher
supervises responses generated by the student under the same staged
progression. Finally, a capability-guided allocation step converts
evaluation feedback into the next training mixture, closing an
evaluation--selection--update loop in which what the system learns to do
shapes what it learns from next.

Evaluated across ten benchmarks spanning harness-based agents, tool
use, coding, and instruction following, this recipe yields three findings.
First, agentic post-training delivers consistent gains at both scales,
lifting the macro-average score of the 4B model from 58.94 to 64.87 and of
the 9B model from 65.60 to 69.04; representative trajectory analyses
illustrate more complete workflows of evidence acquisition, constraint
revision, verification, and artifact delivery, rather than improvements
limited to isolated answers. Second, scale remains beneficial after
post-training: the larger model keeps a clear advantage on tasks requiring
iterative debugging, recovery from execution failures, and state
maintenance over long action sequences. Third, the results provide a preliminary
validation of how the RSI loop closes: the post-trained models return to
the heterogeneous model pool behind the routing harness, while the
post-trained 4B model substantially narrows the aggregate gap to the 9B
base model. Serving requests with these updated checkpoints generates new
trajectories, routing records, and capability feedback under the same
data pipeline, which in turn can seed the training mixture of the next
iteration, establishing an operational basis for extending the loop across
model generations.

These results should be read as an initial attempt at recursive
self-improvement rather than a definitive demonstration. The current
validation concentrates on agentic and coding capabilities, along with
tool use and instruction following, where we observe preliminary but
consistent gains; the broader range of capabilities that the harness
serves has not yet been evaluated, and extending this recipe beyond
them is a clear direction for future work. The results also reflect a
single pass of the evaluation--selection--update loop, so whether the
gains a model realizes from being used in the harness can continue to
accumulate across successive iterations remains to be tested.

Future work follows directly from these limitations. First, we will
iterate the feedback loop across successive model generations, returning
each improved checkpoint to the harness and testing whether the gains from
being used can continue to accumulate as capabilities evolve. Second, the
harness serves a wider range of tasks, execution environments, and models than the
recipes evaluated here, and extending experience collection,
capability-guided allocation, and curriculum design to those regions is a
direct continuation of the data and evaluation thread. Third, the routing
signal can itself be sharpened: turning the recorded
prediction--action--outcome separation into well-calibrated estimates of
difficulty and deficiency---including supervision that trains the router
itself---would let the harness guide not only which data to use and where
to allocate it, but what a model should attempt next. We view \modelname{}
as an initial prototype on this path---evidence that the everyday
operation of a routing harness already supplies both the experience and
the feedback needed to improve the models that run within it, and a
starting point for moving harness-mediated recursive self-improvement
from design to practice.

\clearpage
\appendix
\section*{Appendix}
\section{Contributions}\label{sec:contributions}

Author names are listed in alphabetical order by surname.
\textsuperscript{*} denotes corresponding authors.

\paragraph{Core Contributors.}
Guoliang Cao\textsuperscript{1},
Guohao Dai\textsuperscript{2},
Tianyu Guo\textsuperscript{1},
Kai Han\textsuperscript{1},
Hailin Hu\textsuperscript{1},
Zihan Jiang\textsuperscript{8},
Xiang Kuang\textsuperscript{1},
Boxun Li\textsuperscript{2},
Yulong Li\textsuperscript{1},
Zehua Pei\textsuperscript{1,5},
Yuchuan Tian\textsuperscript{1,4},
Jiamin Wang\textsuperscript{8},
Yu Wang\textsuperscript{3,*},
Yunhe Wang\textsuperscript{1,*},
Yihong Wu\textsuperscript{1},
Haiyang Xu\textsuperscript{2},
Shuo Zhang\textsuperscript{1},
Hang Zhou\textsuperscript{1}

\paragraph{Contributors.}
Siyang Cheng\textsuperscript{1},
Jiayu Fan\textsuperscript{1},
Wei He\textsuperscript{1},
Qingrui Jiao\textsuperscript{1},
Hongguang Li\textsuperscript{1},
Zhiyuan Li\textsuperscript{2},
Runke Liu\textsuperscript{1},
Xi Liu\textsuperscript{1},
Xinchen Liu\textsuperscript{1},
Sinno Jialin Pan\textsuperscript{5},
Yi Ren\textsuperscript{6},
Liuyang Song\textsuperscript{1,4},
Chenyu Wang\textsuperscript{7},
Bei Yu\textsuperscript{5},
Quanlu Zhang\textsuperscript{2},
Xiangyu Zhang\textsuperscript{8},
Mengyu Zheng\textsuperscript{1},
Yingjie Zong\textsuperscript{1}

\paragraph{Affiliations.}
\textsuperscript{1}TokenRhythm Technologies\quad
\textsuperscript{2}Infinigence AI\quad
\textsuperscript{3}Tsinghua University\quad
\textsuperscript{4}Peking University\quad
\textsuperscript{5}The Chinese University of Hong Kong\quad
\textsuperscript{6}Visionplus Capital\quad
\textsuperscript{7}WX Capital\quad
\textsuperscript{8}Alibaba Group

\section{Case Study}\label{sec:case-study}

\subsection*{Case A: Daily Ticket Reporting under Temporal and Audit Constraints}
\paragraph{Prompt and attachments (abridged).}
Produce a Chinese Markdown report for May 18, 2026, for a specified service desk and hotline channel. The agent receives a 64-row ticket export, a JSON state snapshot with key aliases, reporting guidance, and a local-interface description. It must apply exact desk/channel matching, merge aliases, and select each ticket's latest eligible update within the day. The required artifact includes aggregate counts and source-linked audit rows. Next-day updates must not override in-window states; only the designated report file may be changed.

\begingroup
\footnotesize
\setlength{\tabcolsep}{3.0pt}
\renewcommand{\arraystretch}{1.13}
\begin{longtable}{>{\raggedright\arraybackslash}p{0.12\linewidth}>{\raggedright\arraybackslash}p{0.53\linewidth}>{\raggedright\arraybackslash}p{0.28\linewidth}}
\caption{Execution and artifact evidence for Case A.}
\label{tab:case-ticket-paper-style}\\
\toprule
\textbf{Method} & \textbf{Execution \& artifact evidence} & \textbf{Interpretation}\\
\midrule
\endfirsthead
\toprule
\textbf{Method} & \textbf{Execution \& artifact evidence} & \textbf{Interpretation}\\
\midrule
\endhead
\bottomrule
\endfoot
\rowcolor{black!6}
Qwen3.5-9B &
\emph{Execution.} Reads all four attachments, acknowledges errors in its audit table, and repeatedly rewrites the report. It changes the unique-ticket count from 26 to 20 but retains 30 eligible records.
\par\smallskip
\emph{Artifact.} The final Write reports \textcolor{badred}{20 tickets, but status counts sum to 26 and progress counts to 31}. For ticket 3101, it cites in-window R-003 yet assigns reopened/70\%, the values of next-day R-004, while stating that R-004 is excluded. &
Restating the rules and noticing errors does not produce a consistent report. The final content still breaks both aggregate reconciliation and the link between a source record and its reported state.\\
\rowcolor{oursblue!10}
\modelname{}-9B &
\emph{Execution.} Reads the same four attachment paths, separates exclusions by time, desk, and channel, and successfully writes the report.
\par\smallskip
\emph{Artifact.} Reports \textcolor{goodgreen}{50 eligible records and 19 tickets}, matching independent recomputation. It assigns \textcolor{goodgreen}{resolved/100\% to R-003}, excludes R-004, and retains source-row references. Its status-summary table nevertheless contains two priority labels where status labels are required. &
Produces more faithful time-bounded state assignments and a more auditable report. The remaining label errors limit the claim: improved grounding does not establish complete output consistency.\\
\end{longtable}
\endgroup

\paragraph{Summary.}
Qwen3.5-9B retains contradictory totals and assigns a next-day state to an in-window record. \modelname{}-9B reports 50 eligible records and 19 tickets, matching independent recomputation, and excludes the next-day update while retaining source-row references. Its status-summary table still contains two priority labels in place of status labels.

\subsection*{Case B: Implementing a Time-Leakage Auditor from Repository Requirements}
\paragraph{Prompt and attachments (abridged).}
The user suspects that training samples contain features observed after their cutoff time and asks the agent to complete \texttt{app/audit\_leakage.py}: allow a small temporal tolerance, record genuinely late features in the rejection output, and continue exporting clean samples. The WorkBuddy workspace supplies an existing CLI and empty \texttt{audit()} function, \texttt{data/samples.jsonl}, a README, and dependency notes. The README specifies a five-minute tolerance. Clean records must preserve input order and contain \texttt{sample\_id}, \texttt{cutoff\_time}, and \texttt{feature\_count}; rejection output must contain \texttt{summary} and \texttt{rejected}.

\begingroup
\footnotesize
\setlength{\tabcolsep}{3pt}
\renewcommand{\arraystretch}{1.13}
\begin{longtable}{>{\raggedright\arraybackslash}p{0.12\linewidth}>{\raggedright\arraybackslash}p{0.53\linewidth}>{\raggedright\arraybackslash}p{0.28\linewidth}}
\caption{Repository requirements, implementation, and execution artifacts for Case B.}
\label{tab:time-leakage-en}\\
\toprule
\textbf{Method} & \textbf{Execution \& artifact evidence} & \textbf{Interpretation}\\
\midrule
\endfirsthead
\toprule
\textbf{Method} & \textbf{Execution \& artifact evidence} & \textbf{Interpretation}\\
\midrule
\endhead
\bottomrule
\endfoot
\rowcolor{black!6}
Qwen3.5-9B &
\emph{Execution.} Inspects the entry point and sample data, then rewrites the script. No README read appears in the recorded sequence. The implementation introduces \textcolor{leakbadred}{\texttt{TIME\_TOLERANCE\_SECONDS = 2}}, followed by calls to execute the script and read its outputs.
\par\smallskip
\emph{Code and reported artifact.} The final response classifies s1 as leakage: its cutoff is 10:00 and \texttt{f\_pay} occurs at 10:04, within the repository's allowed window. The code also changes the default clean filename to \texttt{clean\_samples.jsonl}, emits \texttt{features} instead of \texttt{feature\_count}, and writes rejections as a top-level list, \textcolor{leakbadred}{omitting the required summary structure}. &
Implements generic timestamp comparison without preserving the repository's business rule and interface. The deviations affect both sample selection and downstream consumption, rather than merely changing presentation.\\
\rowcolor{leakoursblue!10}
\modelname{}-9B &
\emph{Execution.} Reads the entry point, sample data, dependency notes, and README before implementing the auditor. It uses \textcolor{leakgoodgreen}{\texttt{tolerance\_minutes = 5}} and preserves the existing CLI. It then executes the specified command and reads both \texttt{clean.jsonl} and \texttt{leakage.json}; the tool returns a successful exit.
\par\smallskip
\emph{Observed output.} \texttt{clean.jsonl} retains s1 and s3 in order, with feature counts of 2 and 0. \texttt{leakage.json} contains \texttt{summary} and \texttt{rejected}, recording s2's 10:06 feature and s4's next-day feature. \textcolor{leakgoodgreen}{Within-tolerance samples are retained, genuinely late features are recorded, and output follows the specified schema}. &
Translates the documented threshold and data contract into executable code and checks the actual output files. The improvement lies in consistency across requirement discovery, implementation, and delivery compatibility.\\
\end{longtable}
\endgroup

\paragraph{Summary.}
\modelname{}-9B applies the README's five-minute tolerance and preserves the required output schema. Qwen3.5-9B uses a two-second tolerance, rejects the valid sample s1, and changes the output structure.

\subsection*{Case C: Supporting Sustained Two-Player Gomoku Interaction}

\paragraph{Prompt.}
Use HTML to create a simple Gomoku game.

\begin{figure}[htbp]
\centering
\includegraphics[width=0.48\linewidth]{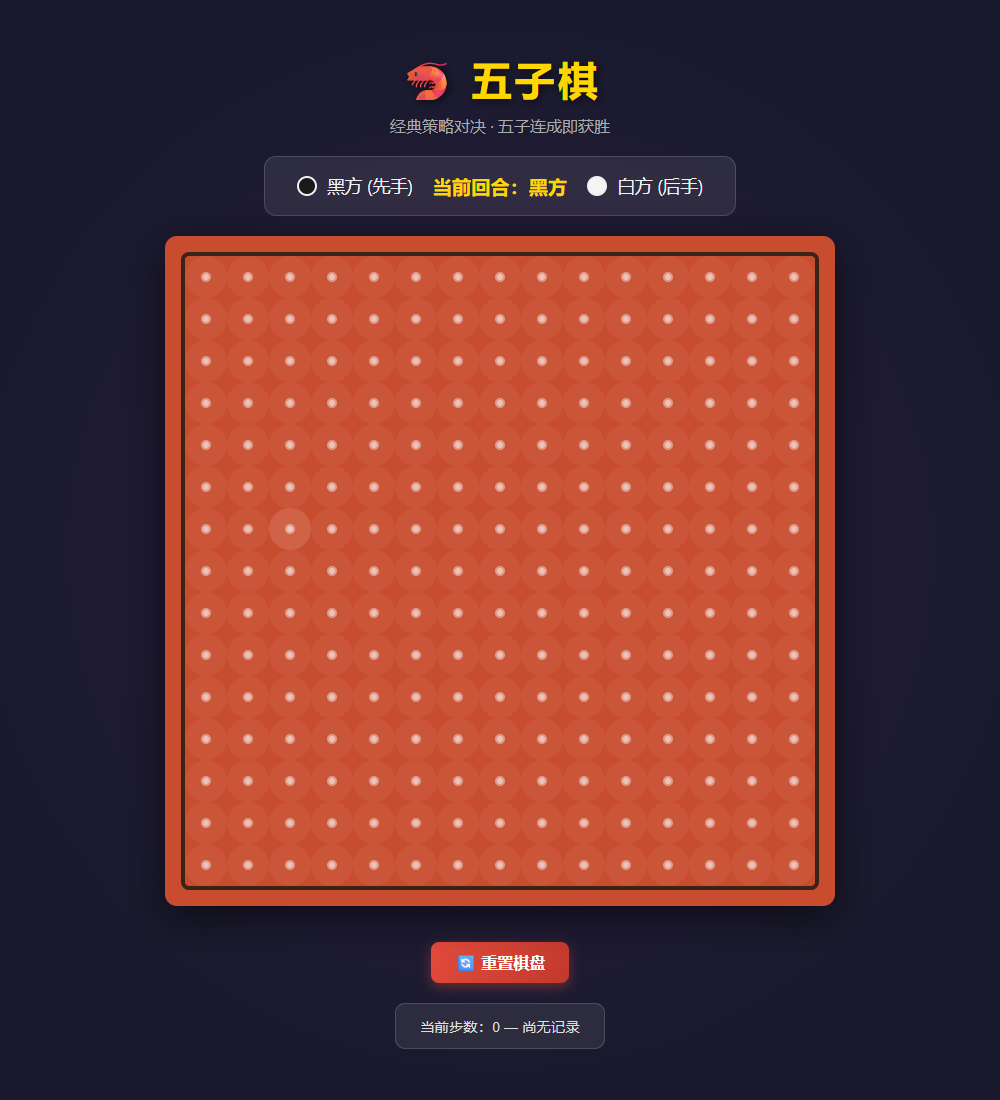}
\hfill
\includegraphics[width=0.48\linewidth]{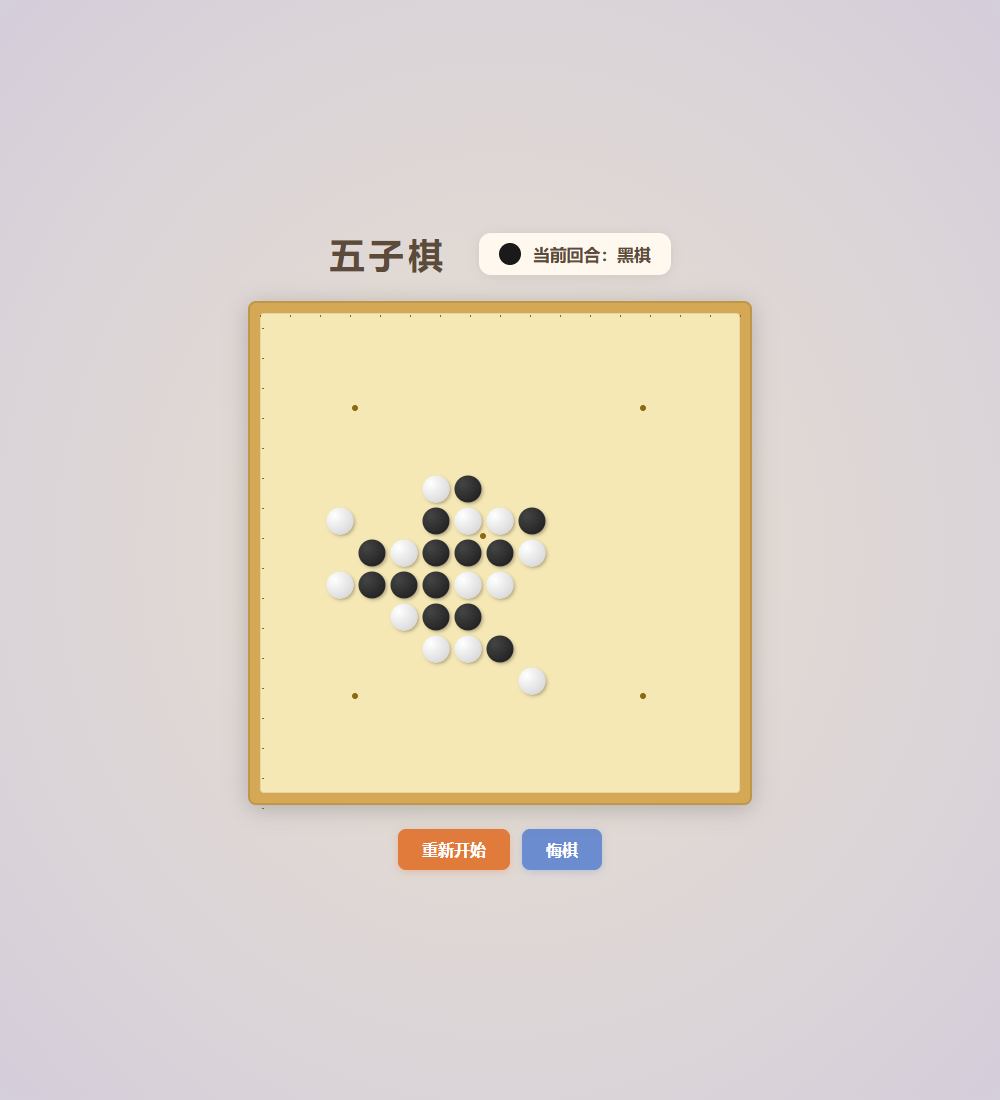}
\caption{Original pages after the same 26-click replay. Left: Qwen3.5-9B remains empty. Right: \modelname{}-9B displays 13 black and 13 white stones in an ongoing position.}
\end{figure}

\begingroup
\small
\setlength{\tabcolsep}{3pt}
\renewcommand{\arraystretch}{1.13}

\begin{longtable}{
    p{0.12\linewidth}
    p{0.50\linewidth}
    p{0.28\linewidth}
}

\caption{Execution and artifact evidence for the Gomoku case.}
\label{tab:gomoku-evidence}\\

\hline
\textbf{Method} &
\textbf{Execution and artifact evidence} &
\textbf{Interpretation} \\
\hline
\endfirsthead

\hline
\textbf{Method} &
\textbf{Execution and artifact evidence} &
\textbf{Interpretation} \\
\hline
\endhead

\hline
\endfoot

\rowcolor{black!6}
Qwen3.5-9B &
\emph{Execution.}
\textcolor{red!70!black}{
All 26 clicks produce indexing errors.
}
The handler reads nonexistent
\texttt{cell.clientX} and \texttt{cell.clientY}
properties from a DOM cell, producing invalid board indices.
\par\smallskip
\emph{Artifact.}
The page renders, but
\textcolor{red!70!black}{
no stones appear and the turn indicator stays on Black.
}
The replay cannot reach normal two-player play.
&
A rendered game interface does not establish working interaction.
Input handling prevents the artifact from accepting and
preserving moves, so the intended attacking and defensive
sequence cannot be executed.
\\

\rowcolor{oursblue!10}
\emph{NeoHorse-1-9B} &
\emph{Execution.}
All 26 clicks execute without captured runtime exceptions.
\textcolor{green!40!black}{
Recorded positions, colors, and visible stone counts
match the input sequence after every move.
}
\par\smallskip
\emph{Artifact.}
\textcolor{green!40!black}{
The page displays 26 stones, preserves both sides'
defensive moves, and returns the turn to Black.
}
The game remains ongoing, with
\texttt{gameOver=false}.
&
Connects input handling, stone rendering, move history,
and turn switching into a usable interaction sequence
along the tested path.
The resulting artifact supports continued play
rather than merely displaying a game-like page.
\\

\end{longtable}
\endgroup

\paragraph{Summary.}
In the tested 26-click sequence, \modelname{}-9B preserves the expected stone positions, colors, and turn order without captured runtime exceptions. Qwen3.5-9B accepts no moves because its click handler produces invalid board indices.

\bibliography{sn-bibliography,neohorse-jev-references}

\end{document}